\documentclass[pdflatex,sn-mathphys]{sn-jnl}

\usepackage[utf8]{inputenc}
\usepackage{amsmath,amssymb}
\usepackage{graphicx}
\usepackage{subfig}
\usepackage{caption}
\usepackage{url}
\usepackage[numbers]{natbib} 
\usepackage{placeins}
\usepackage{xcolor}
\usepackage{siunitx}
\usepackage{url}

\usepackage{xspace}

\usepackage{enumitem}

\usepackage{booktabs} 
\usepackage{multirow}
\usepackage{multicol}
\usepackage{makecell}

\usepackage{etoolbox}

\usepackage{setspace}

\definecolor{wisconsin-red}{rgb}{0.6,0,0}
\definecolor{darkgreen}{rgb}{0.2,0.6,0.2}
\definecolor{maroon}{rgb}{0.5, 0.0, 0.0}
\definecolor{violet}{rgb}{0.75, 0.0, 1.0}
\definecolor{lightgray}{gray}{0.9}
\definecolor{navyblue}{rgb}{0.0, 0.0, 0.5}
\definecolor{darkmidnightblue}{rgb}{0.0, 0.2, 0.4}
\definecolor{midnightblue}{rgb}{0.0,0.4,0.85}
\definecolor{Gray}{gray}{0.75}
\definecolor{darkgreen}{rgb}{0,0.5,0}
\definecolor{apricot}{rgb}{0.98, 0.81, 0.69}

\newcolumntype{C}[1]{>{\centering\arraybackslash}p{#1}}
\newcolumntype{P}[1]{>{\raggedright\arraybackslash}p{#1}}

\begin{document}

\title[Multi-Step Wind Power Forecasting]{A Concurrent CNN-RNN Approach for Multi-Step Wind Power Forecasting}

\author[1]{\fnm{Syed} \sur{Kazmi}}
\author[2]{\fnm{Berk} \sur{Gorgulu}}
\author*[1]{\fnm{Mucahit} \sur{Cevik}}\email{mcevik@torontomu.ca}
\author[3]{\fnm{Mustafa Gokce} \sur{Baydogan}}

\affil[1]{
\orgname{Toronto Metropolitan University}, \orgaddress{\street{44 Gerrard St E}, \city{Toronto}, \postcode{M5B 1G3}, \state{Ontario}, \country{Canada}}}
\affil[2]{
\orgname{University of Toronto}, \orgaddress{\street{ 5 King's College Rd}, \city{Toronto}, \postcode{M5S 3G8}, \state{Ontario}, \country{Canada}}}
\affil[3]{
\orgname{Bogazici University}, \orgaddress{\street{ Bebek}, \city{Istanbul}, \postcode{34342}, \country{Turkey}}}

\abstract{
Wind power forecasting helps with the planning for the power systems by contributing to having a higher level of certainty in decision-making. 
Due to the randomness inherent to meteorological events (e.g., wind speeds), making highly accurate long-term predictions for wind power can be extremely difficult. 
One approach to remedy this challenge is to utilize weather information from multiple points across a geographical grid to obtain a holistic view of the wind patterns, along with temporal information from the previous power outputs of the wind farms. 
Our proposed CNN-RNN architecture combines convolutional neural networks (CNNs) and recurrent neural networks (RNNs) to extract spatial and temporal information from multi-dimensional input data to make day-ahead predictions. 
In this regard, our method incorporates an ultra-wide learning view, combining data from multiple numerical weather prediction models, wind farms, and geographical locations. 
Additionally, we experiment with global forecasting approaches to understand the impact of training the same model over the datasets obtained from multiple different wind farms, and we employ a method where spatial information extracted from convolutional layers is passed to a tree ensemble (e.g., Light Gradient Boosting Machine (LGBM)) instead of fully connected layers. 
The results show that our proposed CNN-RNN architecture outperforms other models such as LGBM, Extra Tree regressor and linear regression when trained globally, but fails to replicate such performance when trained individually on each farm. 
We also observe that passing the spatial information from CNN to LGBM improves its performance, providing further evidence of CNN's spatial feature extraction capabilities. 
}


\maketitle

\section{Introduction}
Rapid economic development and the continuous rise of living standards have raised the need for electric power production in recent years. 
The most common form of energy extraction is from fossil fuels, such as coal, oil, and natural gas. 
However, using fossil fuels comes with serious consequences such as air pollution, ozone depletion and global warming. 
Furthermore, due to the non-renewable nature and limited reserves, unrestrained exploitation of fossil fuels might lead to energy resource depletion. 
According to the Paris agreement, to achieve the goal of limiting the global temperature rise below 2 \textcelsius, renewable energies have to supply two-thirds of the global energy demand up to the year 2050 \citep{hanifi2020critical}. 
The need for a pollution-free and environmentally friendly form of electricity generation has attracted increasing attention over the years and has brought significant focus on renewable sources of energy.

Renewable energy sources such as solar photovoltaic, tidal and modern bioenergy play a crucial role in reducing global carbon footprint by acting as clean alternatives to fossil fuels. 
Wind power generation has witnessed rapid growth over the years for its abundance of availability, low land-based utility, and economic feasibility. 
The wind is a significant and valuable source with the potential to produce energy continuously and sustainably. 
It has the potential to generate electricity for each hour of the day, unlike for example solar energy, which cannot operate at night, and is suitable for systems that require energy continuously. 
Additionally, wind turbines can be built without occupying large areas of land, preventing the loss of agricultural areas. 
Accordingly, wind power systems have developed rapidly around the world as a promising avenue for renewable energy. 
They have become an important component of the smart grid, smart microgrids, smart buildings and smart homes, playing a big role in providing electric power supply.

The use of wind energy has several challenges. 
Due to the intermittent nature of wind and its corresponding environmental factors, wind power production becomes inherently stochastic, which makes grid distribution planning and resource scheduling extremely difficult. 
Additionally, sudden dramatic fluctuations in the wind speed cause the turbines to rotate at a much faster rate than usual, causing sharp increases in electricity production. 
Such an event is referred to as a ramp event. 
On the other hand, when wind speed is too low, the wind turbines do not rotate as fast, leading to a sharp decrease in electricity production, leading to what is known as a down ramp event. 
This often contributes to equipment damage, transmission and distribution losses, and capital loss. 
However, this can effectively be dealt with by employing accurate wind power and ramp event forecasting, which can enable informed and reliable decision making, allowing for better planning, improved efficiency and reduced risk. 
As such, accurate forecasting can play an important role in reducing operating costs and enhancing the competitiveness of wind power systems in the energy industry. 

Wind power forecasting strategies often rely on temporal information extracted from the past production outputs of a wind farm, as well as spatial information derived from meteorological readings at various locations across a geographical grid.
Various wind power forecasting strategies make use of learning techniques to generate accurate predictions. 
While the data for production outputs is specific to the power curve of each wind farm, meteorological data for a given geographical location can be found using Numerical Weather Prediction (NWP) models. 
NWP models provide a complete forecast of the state of the atmosphere at a given time, and a geographical location based on its latitude and longitude coordinates. 
Most wind power forecasting works in recent literature make concurrent use of historic outputs and NWP data and apply learning methods to make reliable short-term and long-term predictions. 
With the advances in artificial intelligence and machine learning (ML) technologies, a large number of deep learning-based models have been considered for wind speed and wind power forecasting due to their superior ability to deal with complex nonlinear problems. 

Wind power forecasting models can be categorized according to their forecast horizons. 
These include ultra short-term forecasts, ranging from a few seconds to 30 minutes ahead, which are useful for turbine control and power load tracking in real-time. 
Short-term forecasts, ranging from 30 minutes to 6 hours ahead, are often used for load dispatch planning. 
Medium-term forecasts, ranging from 6 hours to 1 day, are utilized for energy trading and power system management.
Lastly, long-term forecasts, from 1 day to 1 week or more ahead, allow for optimal maintenance scheduling. 
Our study focuses on medium-term forecasting, with 1-day ahead predictions for both wind power forecasting and ramp detection, and our proposed models take into consideration long-term historical trends (up to 48 hours) as the lookback window. 

\paragraph{Research objectives and contributions}
Our main research objective is to design novel ML models to achieve highly accurate wind power forecasts based on meteorological data and past wind power production output.
We propose a novel multi-head, multi-layer, deep architecture, which combines Recurrent Neural Network (RNN) and Convolutional Neural Network (CNN) structures in parallel to extract spatio-temporal information from meteorological NWP data, and sequential information from historic wind power data. 
We evaluate our model on data collected from seven unique wind farms and compare its performance when trained on each farm independently, against when the model is trained on a combined wind farm data. 
The contributions of this study can be summarized as follows:
\begin{itemize}
    \item We propose a novel CNN-RNN architecture which extracts spatial and temporal information in parallel for improved learning. Our numerical analysis shows that the proposed model is able to outperform competing ML models.
    
    \item Our method incorporates an ultra-wide learning view, combining data from multiple NWP models, wind farms, geographical locations and atmospheric levels. Our analysis points to the benefits of global learning for wind power forecasting tasks.
    
    \item We design a mechanism to employ CNN layers to extract features from the spatial data, and feed those into another machine learning model (e.g., a tree-based ensemble such as Light Gradient Boosting Machine (LGBM)).
    This way, we examine the effectiveness of interdependent learning using spatial features. 
    
\end{itemize}

\paragraph{Organization of the paper}
The remainder of the paper is organized as follows. Section~\ref{sec:lit_review} provides an overview of the relevant studies on wind power forecasting in the literature and their applications across various domains. 
Section~\ref{sec:methodology} introduces our proposed CNN-RNN architecture and the combined CNN and LGBM approach. 
Section~\ref{sec:num_study} provides the experimental setup in terms of evaluation techniques, metrics and model parameters, followed by numerical results. 
Lastly, Section~\ref{sec:conclusions} concludes the paper with a summary of our findings and a discussion on future research directions.

\section{Literature Review}\label{sec:lit_review}

Time series forecasting has been a prominent research field with applications in various domains and it has undergone major methodological advancements over the recent years. 
Earlier studies focused on linear statistical models such as auto-regressive (AR), moving average (MA) and auto-regressive integrated moving average (ARIMA), which account for linear correlations between past data points to make future predictions \citep{box2015time}. 
With the growing availability of exogenous variables, ML models such as Random Forests (RF), Support Vector Machines (SVM) and eXtreme Gradient Boosting (XGB) grew in popularity for their effectiveness in dealing with cross-sectional feature spaces \citep{liu2017towards}. 
More recently, RNNs such as Long Short Term Memory (LSTM)~\citep{hochreiter1997long} and Gated Recurrent Unit (GRU)~\citep{chung2014empirical} architectures have been frequently employed for forecasting tasks due to their ability to extract long-term dependencies between temporal sequences. 
Similarly, CNN-based architectures have been used for time series forecasting due to their ability to capture information along spatial and time coordinates \citep{gu2018recent}.

Various studies pointed to improved forecasting performance when combining multiple methods, which allows for better distinguishing patterns from noise. 
Commonly used ensemble techniques include stacking, bagging and boosting, which have been applied to obtain more accurate forecasting performance than the ones that constitute the ensemble~ \citep{makridakis2020m4}. 
\citet{galacia2019} used decision trees, gradient boosted trees and random forest models for forecasting big data time series, such that the predictions for each ensemble member are obtained by dividing the forecasting problem into forecasting sub-problems.
\citet{makridakis2020m4} conducted a study on the M4 forecasting competition which involves 100,000 time series and 61 forecasting methods. 
They observed improved results when multiple methods were combined to obtain the forecasts.
They noted that using a single model might lead to the difficulty of separating the pattern from the noise. 
Custom boosting algorithms were also shown to achieve high performance for time series modeling~\citep{taherkhani2020adaboost}. 
For instance, \citet{ilic2020b} proposed explainable boosted linear regression, a method which involves training a generic forecasting model to obtain the initial forecasts and then exploring the residuals of the existing model using a regression tree which is trained on all available features.

Recent studies have primarily focused on deep learning models for time series forecasting, with performance improvements over standard approaches for large datasets consisting of a large number of time series~\citep{chen2020probabilistic, rangapuram2018deep, salinas2020deepar}.
Some of these studies adopt a global learning approach to forecasting where training is performed over multiple related time series together in order to capture the seasonal behaviors and dependencies across them.
\citet{hewamalage2022global} demonstrated that no matter how heterogeneous the data may be, a global forecasting model, that can perform equally well, or even better than a collection of independent models always exists.
Complex structures such as DeepAR~\citep{salinas2020deepar}, temporal fusion transformer~\citep{lim2021temporal}, Spacetimeformer~\citep{grigsby2021long}, and N-BEATS~\citep{oreshkin2019n} are examples of models which effectively make use of global learning when provided with large enough samples of related time series. 
These complex deep learning architectures also provide probabilistic forecasting capabilities, for which the typical objective is to predict the parameters of the underlying probability distribution (i.e., mean and variance) for the target value. 
\citet{alexandrov2020gluonts} provided implementations for different probabilistic time series models and created an extensive Python library.

Time-series forecasting for power generation from renewable energy sources is considered to be challenging due to the uncertainties associated with natural events. 
Most energy problems take advantage of long historical patterns of production output along with domain-specific and seasonality-based features to make future predictions. 
Many studies employ ML techniques to effectively forecast for long-term and short-term power generation in order to ensure smooth operational planning and efficient distribution of resources. 
\citet{ozoegwu2019artificial} created a hybrid method based on a combination of nonlinear autoregressive and structural artificial neural networks to forecast monthly mean global solar energy production on a daily basis. 
Similarly \citet{gao2019day} used LSTM networks to make day-ahead solar power generation predictions. \citet{dehghani2019prediction} applied the Grey Wolf optimization method~\citep{mirjalili2014grey} coupled with an adaptive neuro-fuzzy inference to forecast the monthly hydropower generation. 
Other ML models deployed in the energy forecasting domain include SVMs~\citep{sharma2011predicting}, ANNs~\citep{voyant2017machine}, and CNNs~\citep{khosravi2022using}. 

Wind power forecasting is arguably the most challenging form of energy forecasting due to the random fluctuations inherent to wind speeds. 
Previous studies used information from meteorological factors, including wind speed, wind direction, humidity and temperature, recorded at various locations and atmospheric levels across a wind farm, to obtain a diverse set of features for the prediction task.
As such, wind power datasets involve dense spatial attributes, making them a natural fit for CNN-based architectures. 
Below, we discuss previous works within the wind power forecasting domain, which effectively utilize CNNs to extract spatio-temporal information from nonlinear meteorological features.

\citet{yildiz2021improved} proposed a novel residual-based CNN, where historical wind patterns across 54 different wind turbines are concatenated to be represented as 2D RGB images. 
The architecture is able to utilize spatial attention and extract daily and hourly correlations of input data more effectively than other state of the art deep networks including AlexNet~\citep{alippi2018moving}, SqueezeNet~\citep{iandola2016squeezenet}, ResNet-18~\citep{canziani2016analysis}, VGG-16~\citep{alippi2018moving}, and GoogLeNet~\citep{ballester2016performance}. 
\citet{kazutoshi2018feature} proposed a similar 3D-CNN architecture, where wind data from a $50\times 50$ grid is given a video-like representation to account for spatial and temporal information. 
\citet{ju2019model} extracted spatio-temporal information from CNN layers and fed the flattened output to a LGBM model~\citep{ke2017lightgbm}. 
They noted that lack of adequate training data may cause nonlinear convolutional output to fall into a local optimum, which can be avoided by replacing the fully connected layer with a stronger classifier. 
While they feed CNN output to the LGBM model exclusively, our work involves feeding CNN output on top of the original input features. 

Several studies combined the CNN and RNN structures to enhance the prediction performance.
ConvLSTM~\citep{shi2015convolutional} is an example of such combined structures. 
It is an extension of standard LSTM networks which replaces matrix multiplication with convolution operation at each gate in the LSTM cells to capture the underlying spatial features present in multi-dimensional data.
\citet{chen2020short} and \citet{agga2021short} compared 1D and 2D variants of the ConvLSTM network against standalone LSTM and CNN networks. 
\citet{wu2021ultra} proposed a combined CNN-RNN architecture, where spatio-temporal information from multiple meteorological factors of previous timesteps is extracted using CNN and then fed into LSTM to extract long-term historical temporal relationships. 
Alternatively, \citet{zhen2020hybrid} proposed BiLSTM-CNN, where temporal information for each meteorological factor is first extracted using a BiLSTM model, then fed into CNN to extract spatial dependencies. 
They noted that extracting temporal characteristics of input historical sequences first and then feeding them into CNN results in higher prediction accuracy than doing the vice versa. 

Different from these approaches, the AMC-LSTM architecture by \citet{xiong2022short} extracts spatial and temporal features in parallel using CNN and LSTM, respectively, before fusing them together to make final predictions. 
Such parallel structures are computationally inexpensive~\citep{yang2018emotion} and they are included in our study as well. 
Additionally, their architecture uses attention mechanism to effectively assign feature weights based on influence factors. 
\citet{xiang2022ultra} also incorporated attention mechanism within their SATCN-LSTM architecture and they employed a model validation strategy in order to select the best performing version of their model based on validation loss. 
While our work does not include an attention-based mechanism, it does incorporate a similar model validation strategy.

Different from many of the previous works, our analysis is based on a dataset from a massive grid consisting of seven wind farms, each with 48 turbine locations, and 20 unique atmospheric levels. 
Since our dataset is extremely dense, instead of taking into consideration historic wind speed patterns from across all sources, we extract temporal information only from the power curve, as it is a function of all meteorological features, and extract corresponding spatial information at each time step independently. 
A comparative summary of relevant studies is provided in Table~\ref{tab:lit_review}.

\renewcommand{\arraystretch}{1.5} 
\begin{table}[!ht]
\centering
    \caption{Summary of relevant papers in the wind power forecasting domain}
    \label{tab:lit_review}
    \begingroup
\resizebox{0.99\textwidth}{!}{
\begin{tabular}{P{0.09\textwidth} P{0.14\textwidth} P{0.6\textwidth} P{0.08\textwidth} P{0.08\textwidth}P{0.08\textwidth}P{0.08\textwidth}P{0.08\textwidth}P{0.08\textwidth}P{0.12\textwidth}}
\hline
\textbf{Paper} & \textbf{Architecture} & \textbf{Methodology} & Forecast horizon & Temporal resolution & Data instances & \# locations & \# wind farms & \# atmospheric levels\\
\hline
\hline
\citet{yildiz2021improved} & ResCNN & 2D-CNN to extract spatio-temporal information & 1, 2, 3 step & 1 hr & 70080 & 54 & 1 & 2 \\ 
\hline
\citet{kazutoshi2018feature} & 3D-CNN & 3D-CNN to extract spatio-temporal information & 96 step & 30 min & 40320 & 50x50 grid & 1 & 2 \\ 
\hline
\citet{jiajun2020ultra} & WT-DBN-LGBM & Features extracted using DBN fed to LGBM & 1, 2, 3 step & 10 min & 800000 & 4 & 1 & 1 \\ 
\hline
\citet{ju2019model} & CNN-LGBM & Spatio-temporal information extracted using CNN fed to LGBM & 1 step & 5 min & - & 5 & 1 & 1 \\ 
\hline
\citet{chen2020short} & ConvLSTM1D & ConvLSTM1D to extract temporal information from univariate time series & 1 step & 15 min & 6000 & 3 & 1 & 1 \\ 
\hline
\citet{agga2021short} & ConvLSTM2D & ConvLSTM2D to extract spatio-temporal information from multivariate time series & 1, 3, 5, 7 step & 24 hr & - & - & 1 & 1 \\ 
\hline
\citet{wu2021ultra} & STCM & Spatio-temporal information from previous timesteps extracted using CNN and then fed into LSTM & 12 step & 5 min & 104800 & 33 & 1 & 1 \\ 
\hline
\citet{zhen2020hybrid} & BiLSTM-CNN & Temporal information from previous timesteps extracted using BiLSTM and then fed into CNN & 1 step & 5 min & 4896 & - & 1 & 4 \\ 
\hline
\citet{xiang2022ultra} & SATCN-LSTM & Spatio-temporal information from previous timesteps extracted using CNN and then fed into LSTM using attention mechanism & 16 step & 5 min & 10468 & 16 & 2 & 1 \\ 
\hline
\citet{xiong2022short} & AMC-LSTM & Spatio-temporal information from previous timesteps extracted using CNN fed into LSTM, temporal  information from previous wind power timesteps extracted using LSTM & 1, 2, 3, 5 step & 3 min & 13440 & 1 & 1 & 1 \\ 
\hline
Our study & CNN-RNN & Spatial information for each future timestep extracted using CNN, temporal information from previous wind power timesteps extracted using LSTM & 24 step & 1 hr & ~ 12000 & 48 & 7 & 20 \\ 
\hline
\end{tabular}
}
\endgroup
\end{table}

\section{Methodology}\label{sec:methodology}
In this section, we discuss the various strategies used for time series forecasting. 
We first provide details on our dataset, including the distribution of the production outputs as well as general data characteristics.
Then, we elaborate on the models and architectures employed in our analysis and assess their strengths and drawbacks.
Finally, we provide our proposed architectures for time series forecasting.

\subsection{Dataset}
In our analysis, we use meteorological data from seven wind farms located in Turkey, which was extracted using the Global Forecast System (GFS), and the Action de Recherche pour la Petite Echelle et la Grande Echelle (ARPEGE) NWP model, with meteorological features compromising Pressure (Pa), Temperature (K), Relative Humidity (\%), and vertical (VGRD) and horizontal (UGRD) components of wind speed (m/s). 
Our analysis encompasses only a normalized vector of UGRD and VGRD as the unique meteorological feature, which we refer to as wind speed. 
Data for each farm was taken from early 2020, up until March 2022, with the sample size averaging around 12,000 data points. 
A summary of further data characteristics of GFS and ARPEGE datasets is provided in Table~\ref{tab:gfs_arpege_comparison} below.  

\setlength{\tabcolsep}{7.5pt} 
\renewcommand{\arraystretch}{1.5} 
\begin{table}[!ht]
\centering
\caption{Comparison of GFS and ARPEGE data characteristics}
\label{tab:gfs_arpege_comparison}
\begingroup
\resizebox{0.60\textwidth}{!}{
\begin{tabular}{lll}
\hline
 & \textbf{GFS} & \textbf{ARPEGE} \\ 
\hline
Temporal Resolution & 3-hourly & hourly \\
Height Levels & 24 & 27 \\
Latitudes & 4 & 5 \\
Longitudes & 4 & 5 \\ 
\hline
\end{tabular}
}
\endgroup
\end{table}

After carefully analyzing the correlations between the different atmospheric levels and wind power, we selected atmospheric level features that were most relevant to power prediction. 
This allowed us to select 11 unique levels from ARPEGE data and 9 unique levels from GFS data. 
Since the data resolution of GFS 3-hourly instead of hourly, we use the mean of the one-step lag and one-step ahead values to augment the missing values. 
We normalize the wind power production outputs using min-max normalization as shown in Equation~\ref{eq:min_max}. 
In addition to meteorological features, we use cyclic month-of the year (moy) and hour-of-day (hod) features \citep{ilic2020augmented}, which are incorporated using sinusoidal and cosinusoidal transformations as follows:
\begin{align}
    \label{eq:min_max} \hat{x}_{i} = \frac{x_{i} - x_{\min}}{x_{\max} - x_{\min}}
\end{align}
\begin{align} 
\text{month of year}_{\sin} &= \sin(moy \times \frac{2\pi}{7}) & \text{month of year}_{\cos} = \cos(moy \times \frac{2\pi}{7} ) \label{eq: month} \\
\text{hour of day}_{\sin} &= \sin (hod \times \frac{2\pi}{24}) & \text{hour of day}_{\cos} = \cos (hod \times \frac{2\pi}{24} )\label{eq: hour}
\end{align}

\begin{figure}[!ht]
    \centering
    \subfloat[Sample wind power output from 2018] {{\includegraphics[width=0.79\textwidth]{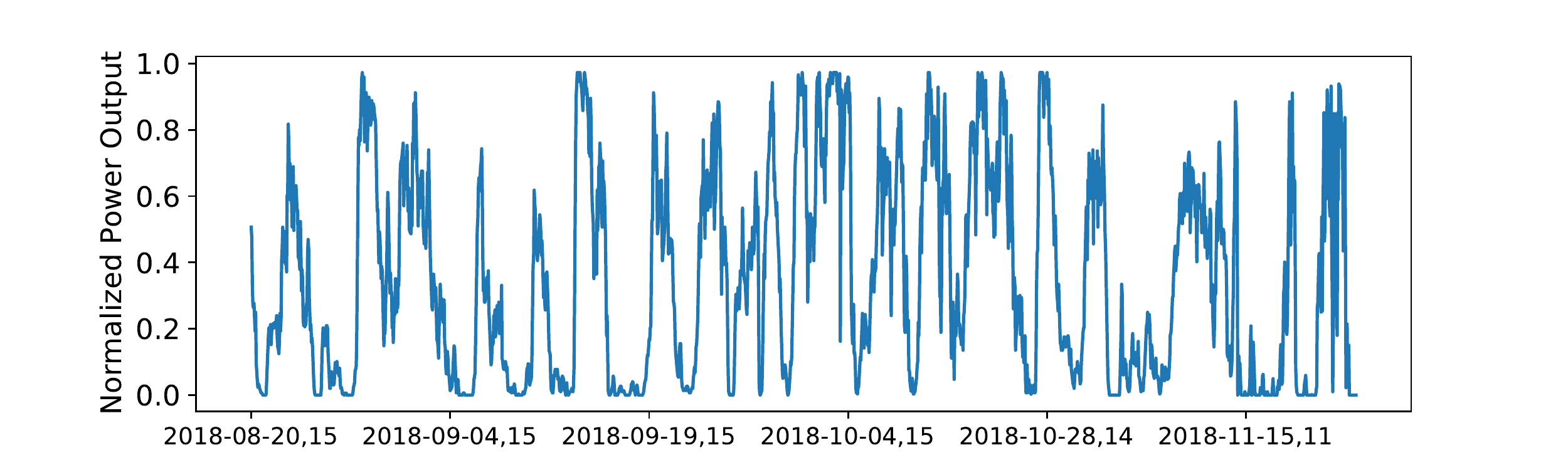}}}\hspace{0.11in}
     \subfloat[Sample wind power output from 2019] {{\includegraphics[width=0.79\textwidth]{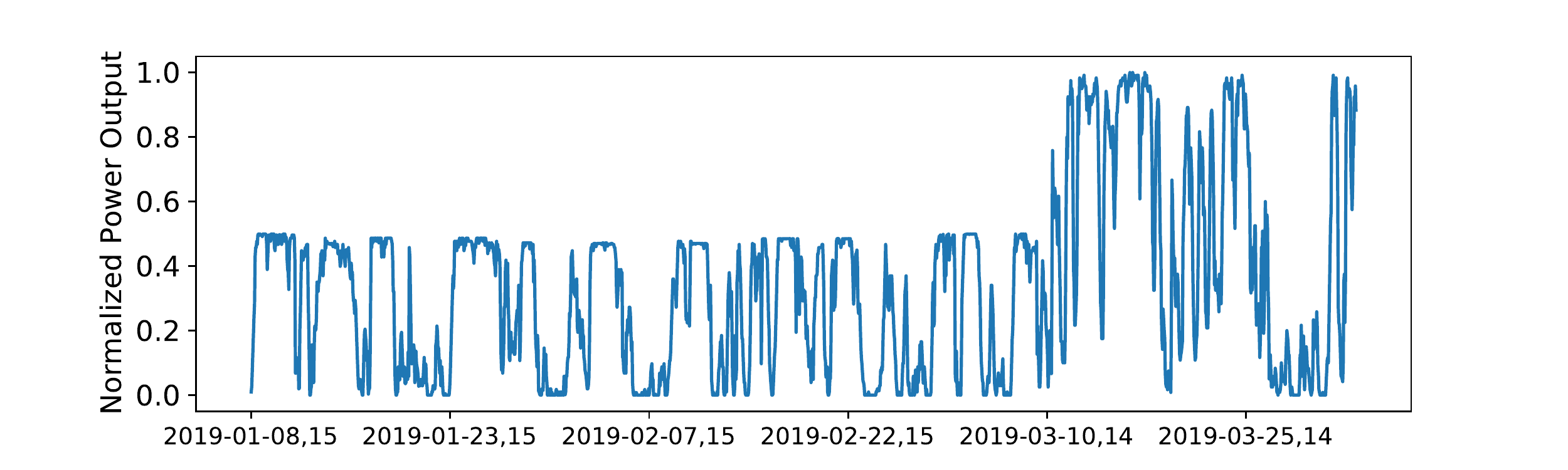} }}\hspace{0.11in}
     \subfloat[Sample wind power output from 2020] {{\includegraphics[width=0.79\textwidth]{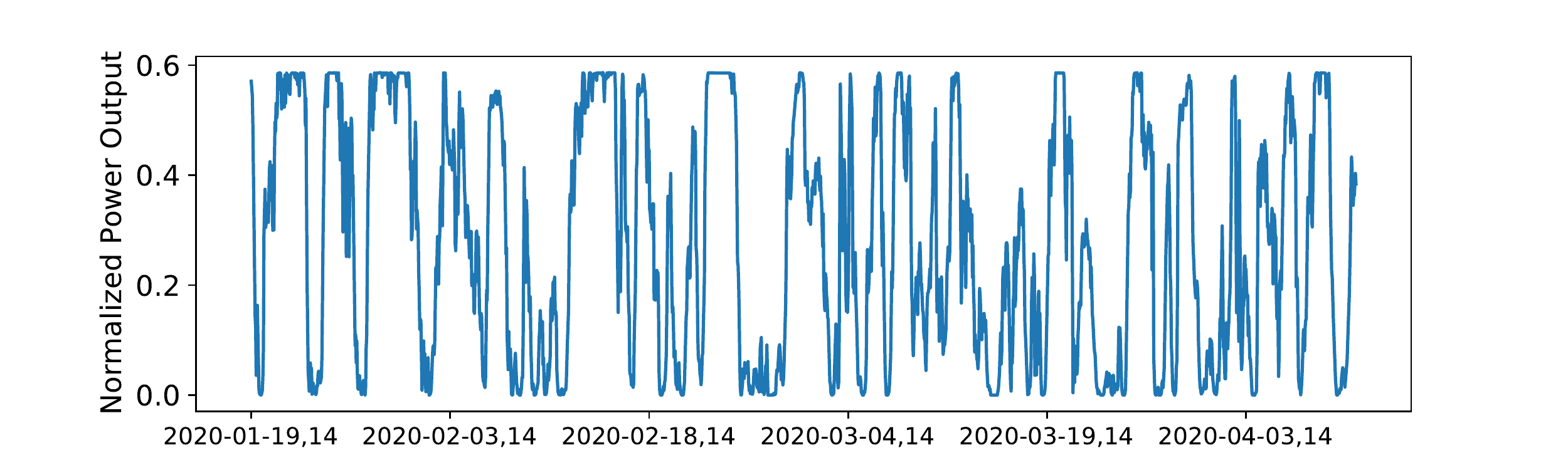}}}
    \caption{Power production output samples from 2018, 2019 and 2020}
    \label{fig:power}
\end{figure}

\begin{figure}[!ht]
    \centering
    \subfloat[Average hourly values \label{fig:data_hourly}] {{\includegraphics[width=0.45\textwidth]{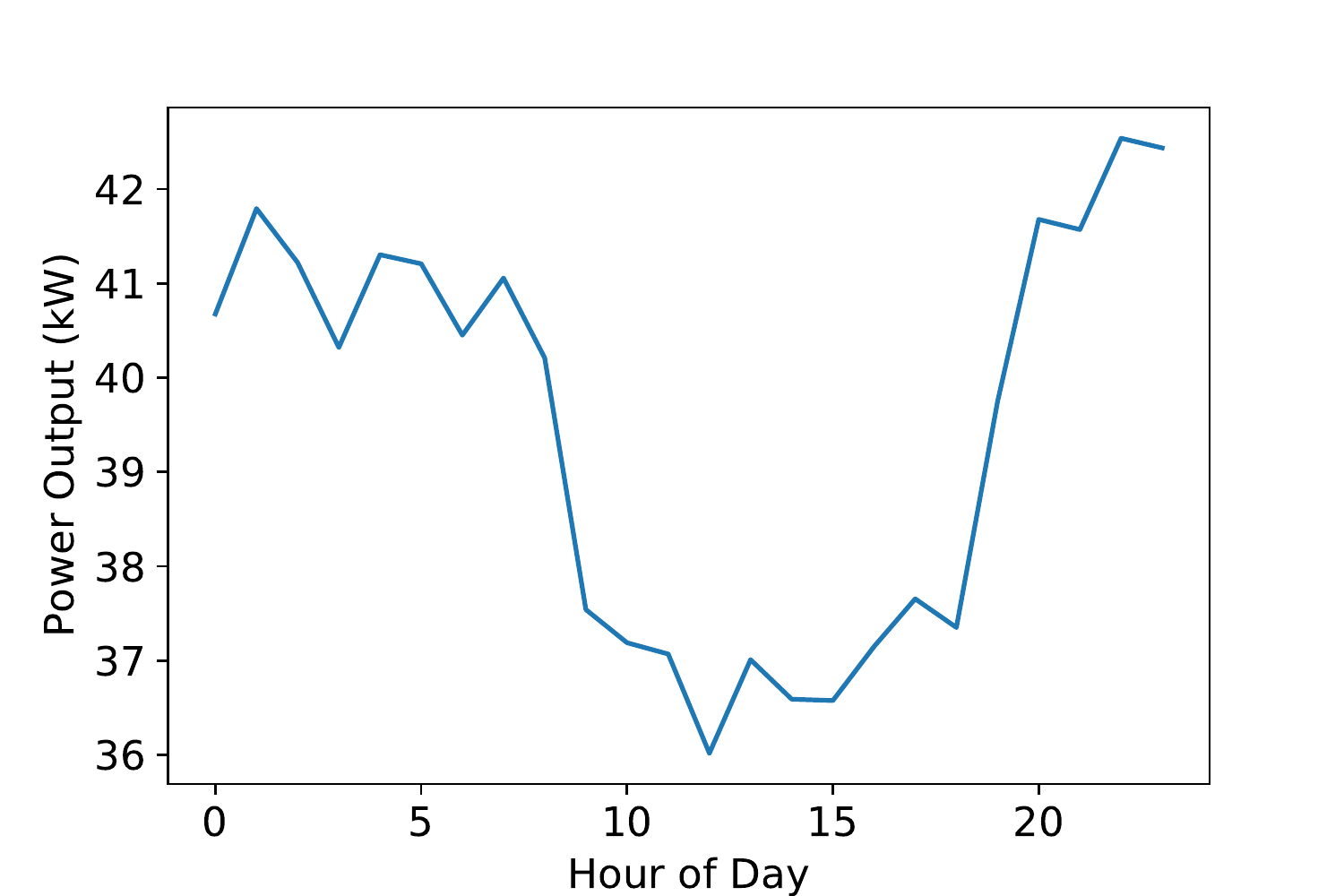}}}\hspace{0.01in}
     \subfloat[Average monthly values \label{fig:data_monthly}] {{\includegraphics[width=0.45\textwidth]{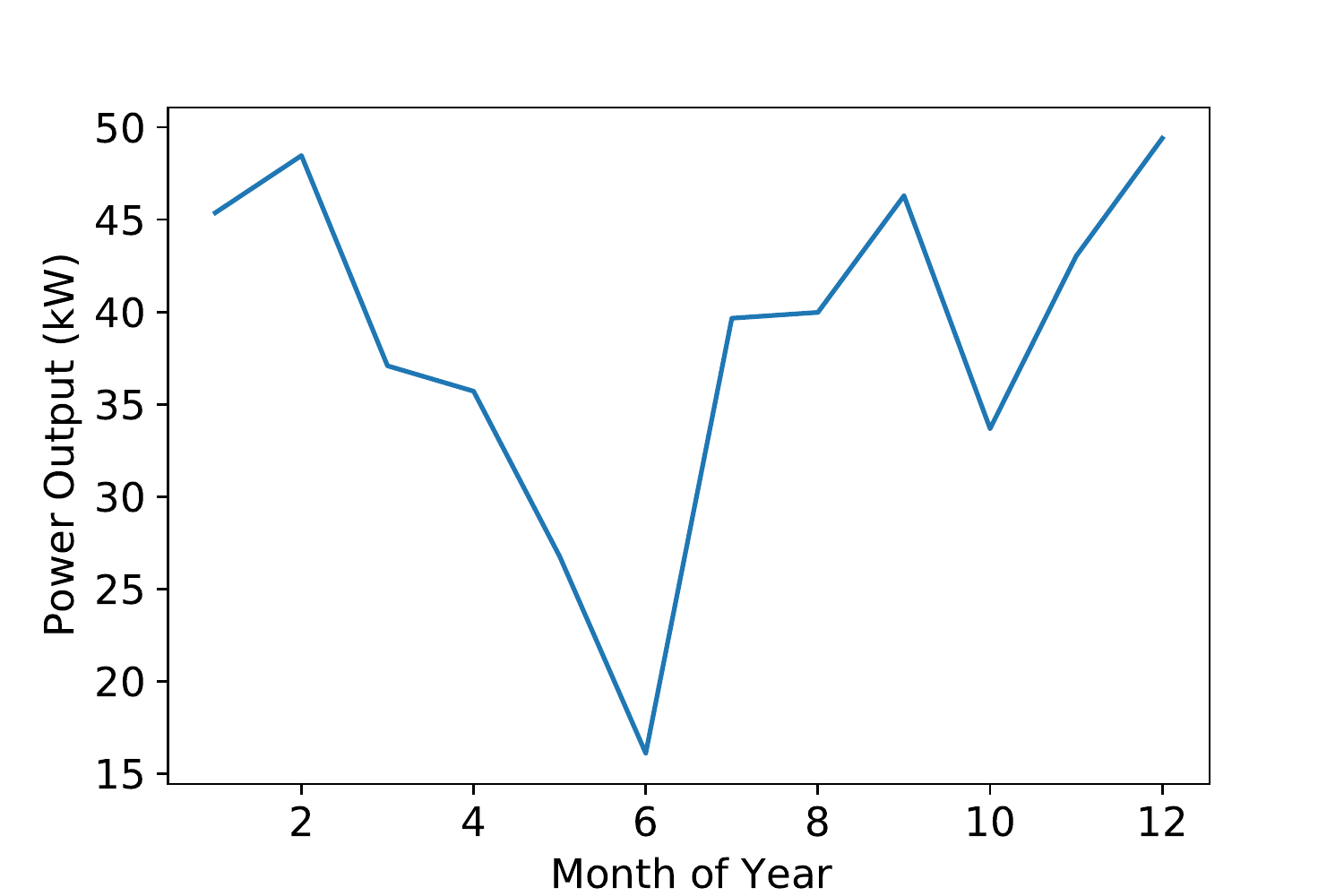}}}
    \caption{Average hourly and monthly trends of power production output}
    \label{fig:hourly_monthly}
\end{figure}

Figure~\ref{fig:power} shows sample normalized wind power output time series from the years 2018, 2019 and 2020 to illustrate the evolution of the wind power data.
Figure~\ref{fig:hourly_monthly} shows hourly and monthly values for the wind power output.
Through the years 2018, 2019 and 2020, the power output distribution is very similar and no substantial yearly changes are observed. 
The hour of the day trend (Figure~\ref{fig:data_hourly}) reveals that power output is maximized after midnight, and the lowest values are observed between noon to around 4 PM. 
Monthly trends (Figure~\ref{fig:data_monthly}) show that production is low during the summer months, with the lowest production in June, and peaks are achieved during winter, especially in January and February. 

\begin{figure}[!ht]
\centering
\includegraphics[width=0.65\textwidth]{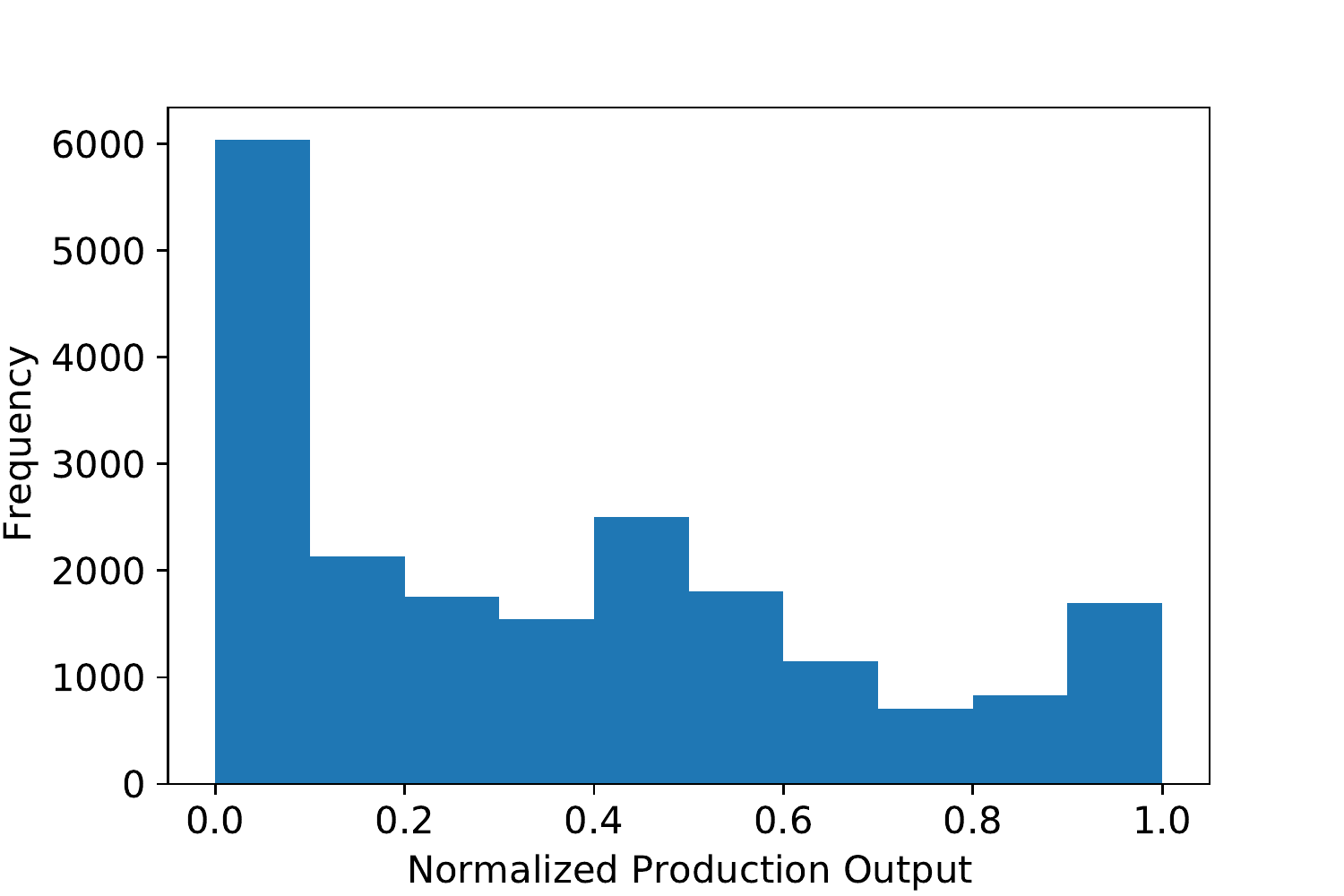}
\caption{Distribution of the production output across all wind farms}
\label{fig:hist_prod}
\end{figure}

Figure~\ref{fig:hist_prod} demonstrates the normalized distribution of the production output.
We note that the output is skewed towards the left, illustrating that most of the production outputs lie within the 10th percentile.
In terms of the ramp events present in the dataset, we note that the ratio of a ramp event against a no-ramp event is 1:5,673. 
This indicates a high class imbalance issue for the ramp detection task, which requires data balancing for model training, e.g., assigning more weight to the class that occurs less frequently. 
On the other hand, undersampling and oversampling cannot be applied to cure the data imbalance due to the nature of time series datasets, removing or adding any additional points to the data disrupts the overall pattern and trend of the time series. 

When combining data for the seven wind farms for global training purposes, we concatenate the independent datasets along the zeroth axis (i.e., concatenating the rows) and label the farms with one-hot encoding.

\subsection{Review of Standard ML Methods for Forecasting}
While several different ML models have been previously used for time series forecasting tasks, in our analysis, we use linear regression as a baseline model, and tree ensembles such as LGBM and extra-trees (ET) as strong baselines. 
Note that we choose these models as they reportedly show high performance for various time series forecasting tasks~\citep{ilic2020b, parmezan2019}.
ML models require careful feature extraction for training high-performance forecasting models.
Commonly extracted features for time series forecasting include features obtained from timestamps, e.g., time of the day, day of the week, and month of the year. 
One-hot encoding or sine/cosine transformation can also be considered for these features.
For our dataset, since we have a linear representation of the features, where at each time step there exists a wind speed value extracted from various locations and height levels, we column-wise concatenate all the features.
Note that the lag values (e.g., the values associated with previous time steps) can also be included in the feature space in a similar manner to allow for a cross-sectional feature matrix. 
However, the main downside of using standard ML models for time series forecasting is that they do not consider the sequential information present in historical patterns. 

In general, forecasting can be done for making one-step or multi-step ahead predictions. 
For one-step-ahead predictions, a base model is trained such that the lag values up to time $t$, along with any exogenous features are used to predict the value for time $t+1$. 
For multi-step ahead forecasting, two strategies are commonly used. 
The first strategy is \textit{Direct Multi-step Forecast Strategy}, which uses a base model to forecast for every time step in a time series. 
For example, in a scenario which requires making $n$-step ahead predictions, a different base model is trained for each of those $n$th step predictions. 
The advantage of this method is that since it uses lags on the same time instance in a particular data series, it is easy to implement and experiment with. 
However, a downside to this approach would be the high computational cost of training each of the separate base models. 
The second approach is to use a multi-output prediction model that is capable of generating multiple predictions. 
These models can learn the dependencies between inputs and outputs as well as those between outputs.
Deep learning models (e.g., RNN- and CNN-based architectures) are typically designed to generate multi-output predictions.

\subsection{Proposed Architectures}
RNNs and CNNs have been commonly employed for time series modeling~\citep{montero2021principles}.
RNNs process given information incrementally, while maintaining an internal model of what is being processed based on the past information, and constantly updating its state as new information is received. 
As such, they are suitable for problems where the sequence of the data matters. 
Specific RNN architectures such as Long Short-Term Memory (LSTM) and Gated Recurrent Unit (GRU) networks are designed to model temporal sequences and their long-range dependencies more accurately than conventional RNNs, and they are frequently employed to capture temporal information in complex neural networks used for time series forecasting~\citep{rangapuram2018deep, salinas2020deepar}.
On the other hand, CNNs are known for their feature extraction capabilities from large datasets~\citep{fawaz2019deep}.
For time series modeling, a CNN can be seen as applying and sliding a filter over the sequential data. 
Unlike RNNs, the same convolution is used to find the relevant values for all the time stamps, which is a powerful property of the CNNs, referred to as weight sharing, as it enables learning filters that are invariant across the time dimension.

In this study, we propose a novel CNN-RNN architecture for the wind power forecasting task, and compare it against standard ML models and vanilla CNN architectures.
In addition, we provide a method, Conv2D + LGBM, which use 2D CNNs as a feature extractor for other ML models.
In this method, we consider LGBM as a representative ML model, however, other ML models such as XGB and Random Forests can be employed to utilize the extracted features from CNNs.

\subsubsection{CNN-RNN}
Our proposed CNN-RNN architecture combines CNN and RNN architectures as follows. 
Two CNN models, one for GFS and the other for the ARPEGE dataset, extract spatio-meteorological information at each time step of the forecast horizon.
Then, final predictions are obtained by combining CNN outputs with the output from an RNN encoder for that time step, which stores temporal information of the previous wind power values.
The idea behind relying solely on the previous wind power values to extract temporal information, while disregarding the historical meteorological features, is that since the wind power output is a function of those features, it summarizes the relevant information contained within the historic meteorological patterns \citep{dorado2020multi}. 
Note that this approach is computationally efficient and it prevents high memory consumption caused by storing information of the previous $n$ time steps, consisting of dense meteorological features accumulated from numerous different sources. 
Additionally, we only select wind speed as the meteorological feature for our analysis, since the goal for our model is to only capture location-wise interdependencies, and not waste computational resources by accounting for interdependencies between the meteorological features and the values at different time steps. 
Since our architecture does not store any previous exogenous features, we do not require using an attention mechanism to pay attention to important segments in a long historical sequence.

\begin{figure}[!ht] 
    \centering
    \includegraphics[width = 0.90 \linewidth]{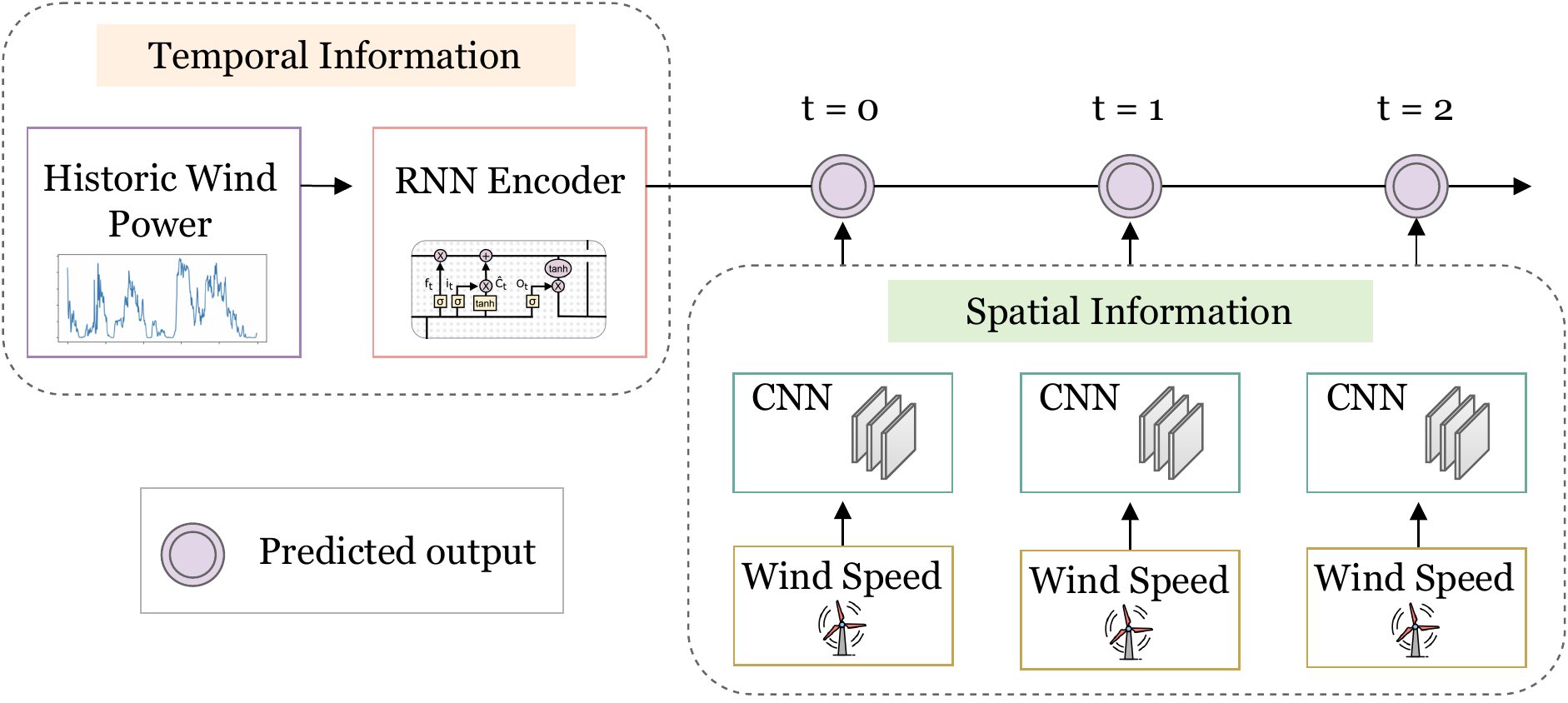}
    \caption{CNN-RNN architecture illustrating a representative three-step ahead prediction.}
    \label{fig:cnn_rnn_architecture}
\end{figure}

In our proposed architecture, historic wind power data from time $t_{-n}$ to $t_{-1}$ is fed into an RNN encoder, which encapsulates the sequential information of the input vector within its internal states $h_t$ (hidden state) and $c_t$ (cell state).
The internal states are then passed along the forecasting horizon in the form of a context vector, where at each time step $t_{+1}$, RNN cell outputs are combined with the flattened output of CNN, generated based on wind speed inputs corresponding to that same particular time step. 
At each time step, CNN models perform two-layered convolution with a filter bank to produce a set of feature maps for the input data, which are then batch normalized.
Then an element-wise ReLU non-linearity, $\max(0,x)$, is applied. 
Following that, max-pooling with a $2 \times 2$ window and stride 2 is performed, and a dropout layer is applied to the resulting output. 
Max-pooling is used to achieve pattern invariance over small spatial shifts in the 2D feature space. 
The combined CNN and RNN output are then passed to a fully connected network involving dense layers, before undergoing a linear activation function to generate the final prediction. 
The proposed approach is summarized in Figure~\ref{fig:cnn_rnn_architecture}.  

Since our proposed method only considers exogenous features for their respective time step, it involves preparing data so that data for all future prediction time steps gets stored in an array. 
This leads to three input arrays in total; one consists of the lag values of length of the considered historical period, and the other two compromising meteorological features of lengths equivalent to the forecasting horizon. 
Hence, the shape of the input tensors for CNN is (\textit{samples}, \textit{timesteps}, \textit{height}, \textit{width}, \textit{channels}), while for the RNN encoder the input is of the form (\textit{samples}, \textit{timesteps}, \textit{feature}). 

\subsubsection{Conv2D + LGBM}
CNNs are well-known for their ability to automatically extract important features from large datasets.
In this regard, the features extracted from CNNs can be fed into another ML model e.g., Random Forests and LGBM to benefit from the strengths of another model to enhance the prediction performance.
For our analysis, we employ LGBM ~\citep{ke2017lightgbm} as the representative ML model, since it consistently provides high performance for our forecasting tasks (as observed in our preliminary analysis) and train LGBM using both the original input features and the extracted features from 2D CNNs (i.e., Conv2D).

\begin{figure}[!ht]
    \centering
    \subfloat[2D spatial data] {{\includegraphics[width=0.35\textwidth]{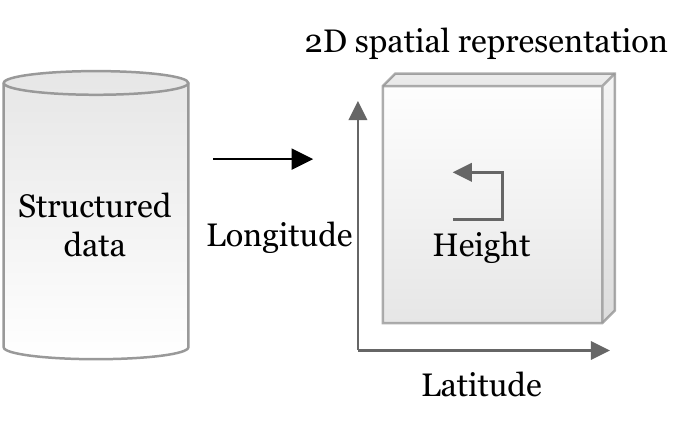}}}\hspace{0.55in}
     \subfloat[Feature extraction by CNN] {{\includegraphics[width=0.40\textwidth]{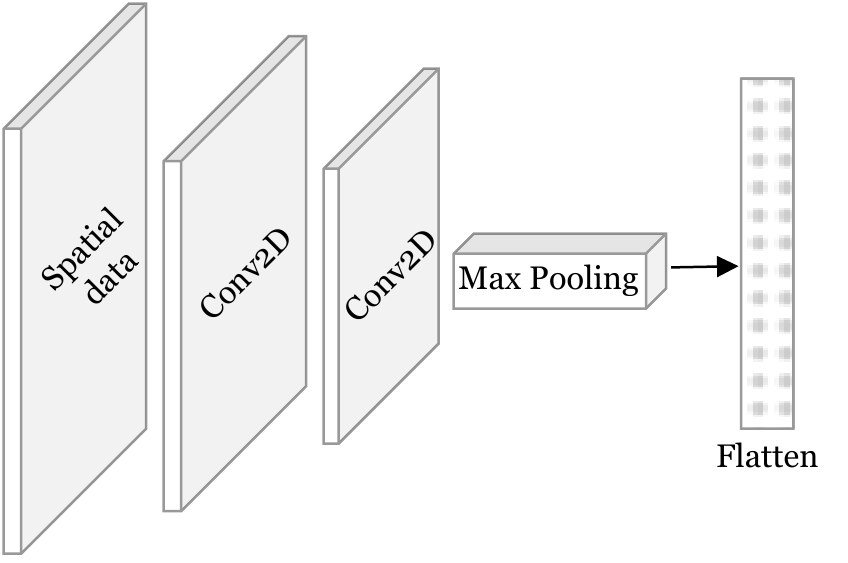} \label{fig:conv_structureb}}}
    \caption{Visual representation of the 2D spatial data; and proposed CNN architecture to extract information from the 2D spatial data}
    \label{fig:conv_structure}
\end{figure}

Our Conv2D + LGBM method makes use of the spatial feature handling capabilities of the initial Conv2D layers within a CNN architecture, while replacing the fully connected layers with a strong LGBM regressor. 
We use two Conv2D layers, followed by a max pooling layer to extract spatial information from the input data, and then pass the flattened output to an LGBM model, which combines this information, along with the original set of input features to train itself. 
The process of retrieving the flattened output is demonstrated in Figure~\ref{fig:conv_structure}. 
This approach might benefit the conventional CNN since replacing the fully connected layers with LGBM can help avoid falling into a local optima due to the amount or quality of the data. 
Similarly, the performance of a traditional LGBM model can be enhanced by this approach, as it provides the LGBM with nonlinear feature interactions that it might not be able to learn otherwise. 

\section{Numerical Study}\label{sec:num_study}
In this chapter, we investigate the effectiveness of different machine learning methods for our wind power forecasting task.
Below, we first provide the details of the experimental setup.
Then, we present the results from our detailed numerical study and discuss the performance improvements that can be attributed to our proposed methods.

\subsection{Experimental Setup}
We first provide the details of our experimental settings including the metrics to evaluate forecasting models, hyperparameters for these models and train-test split of the dataset.
Specifically, in our numerical study, we include four different forecasting methods and seven distinct wind farm datasets to ensure valid research outcomes. 
We conduct the numerical experiments using Scikit-learn version 1.0.2 and TensorFlow version 2.8.2, on a 2.7 GHz dual-core i5 processor with 8GB of RAM.
All the implementations are done in the Python programming language.

\subsubsection{Evaluation metrics}
We consider two performance evaluation metrics, Normalized Deviation (ND) and  Normalized Root Mean Squared Error (NRMSE), to compare the performances of forecasting methods.
\begin{align*}
    \text{ND}(y, \hat{y}) = \frac{\displaystyle \sum_{i=1}^{N}\vert\hat{y_i} - y_i\vert}{\displaystyle \sum_{i=1}^{N} \vert y_i \vert}, \qquad
    \text{NRMSE}(y, \hat{y}) = \frac{\displaystyle \sqrt{\frac{1}{N} \sum_{i=1}^{N}(\hat{y_i} - y_i)^2}}{\displaystyle \frac{1}{N} \sum_{i=1}^{N} \vert y_i \vert}
\end{align*}

\noindent where $y = [y_1, \hdots, y_N]$ and $\hat{y} = [\hat{y}_1, \hdots, \hat{y}_N]$ represent ground truth and predicted values over a prediction horizon $N$, respectively. 
RMSE is a popular metric for assessing the performance of regression models and it is typically the preferred method when the model errors follow a Gaussian distribution \citep{ballester2016performance}. 
ND, likewise, increases linearly with an increase in deviations from the ground truth. 
These metrics are applied to each batch in the test set independently, and the average across the batches is reported as the final performance value.
Statistical significance of the results is measured using the two-sided paired t-test \citep{hsu2014paired} at 95\% as the significance level. 
For ramp classification, we use ``precision'' (a measure of the percentage of instances predicted as ramp actually belongs to the same class), ``recall'' a measure of the percentage of instances detected as ramp are identified correctly), and ``F1-score'', which is the harmonic mean of precision and recall.

\subsubsection{Model settings}

Our proposed architecture is compared against three baseline models, namely linear regression (LR), extra tree regressor (ET), and LGBM. 
Our experiments are conducted using random seed and random state values to mitigate the stochasticity involved with ML model training. 
We perform extensive hyperparameter tuning for all the forecasting models, where, for our deep architectures, we experiment with different combinations of stacked layers and hidden units, along with other parameters including optimizer and batch size. 
For other ML models, we apply grid search to identify the best-performing parameters. 
The final set of hyperparameters used for each model is provided in Table~\ref{tab:hyper_para}, while the search space compromising all different parameter combinations used in our hyperparameter tuning experiments is provided in Table~\ref{tab:hyper_parb}.

\begin{table}[!ht]
\centering
\caption{The hyperparameter settings used in the experiments for the employed models}
\label{tab:hyper_para}
\begingroup
\setlength{\tabcolsep}{7.5pt} 
\renewcommand{\arraystretch}{2.1} 
\resizebox{0.70\textwidth}{!}{
\begin{tabular}{lll}
\toprule
\textbf{Model} &  & \textbf{Final Parameters} \\
\midrule
\multirow[t]{2}{*}{CNN} & & \makecell[l]{\textit{hidden units}: \{264, 128\}, \textit{kernel size}: \{4, 2\}, \textit{strides} : \{1, 1\}, \\*[0.2em] 
\textit{optimizer}: Adam, \textit{loss}: mse, \textit{batch size}: 64, \textit{lookback}: 48
}\\
\midrule
\multirow[t]{2}{*}{LSTM} & & \makecell[l]{\textit{hidden units}: \{128, 64\}, \textit{optimizer}: Adam, \textit{loss}: mse,  \\*[0.2em]
\textit{batch size}: 64, \textit{lookback}: 48
}\\
\midrule
\multirow[t]{2}{*}{ET} & & \makecell[l]{\textit{\# of trees}: 120,  \textit{n\_estimators}: 100 \textit{splitting criterion}: mse,\\*[0.2em] 
\textit{max depth}: $\infty$,  \textit{lookback}: 24
}\\
\midrule
\multirow[t]{2}{*}{LGBM} & & \makecell[l]{\textit{\# of leaves}: 90, \textit{n\_estimators}: 100,  \textit{learning rate}: 0.07,\\*[0.2em]  
\textit{max depth}: $\infty$,  \textit{n\_estimators}:100,  \textit{min\_child\_samples}: 22
}\\
\bottomrule
\end{tabular}
}
\endgroup
\end{table}

\setlength{\tabcolsep}{7.5pt} 
\renewcommand{\arraystretch}{1.17} 
\begin{table}[!ht]
\centering
\caption{Hyperparameter search space for forecasting models}
\label{tab:hyper_parb}
\subfloat[Deep learning models \label{tab:hyper_parb1}]{
\resizebox{0.439\textwidth}{!}{
\begin{tabular}{ll}
\toprule
\textbf{Parameter} & \textbf{Search space} \\ 
\midrule
hidden layers & {[}1, 2, 3, 4{]} \\
hidden units & {[}32, 64, 128, 264, 518{]} \\
kernel size & {[}1, 2, 3, 4, 5, 6, 7, 8, 9{]} \\
strides & {[}1, 2, 3, 4{]} \\
optimizers & {[}Adam, Adamax, SGD{]} \\
batch size & {[}32, 64, 128, 264, 518{]} \\
\bottomrule
\end{tabular}
}
}
\-\hspace{0.3em}
\subfloat[Tree-based models \label{tab:hyper_parb2}]{
\resizebox{0.439\textwidth}{!}{
\begin{tabular}{ll}
\toprule
\textbf{Parameter} & \textbf{Search space} \\ 
\midrule
\# of trees & {[}80, 100, 120, 140{]} \\
n\_estimators & {[}80, 100, 120, 140{]} \\
\# of leaves & {[}30, 60, 90, 120{]} \\
max\_depth & {[}50, 100, 500,  $\infty${]} \\
splitting criterion & {[}mae, mse{]} \\
min\_child\_samples & {[}10:50{]} \\
\bottomrule
\end{tabular}
\label{tab:hyper_para2}
}
}
\end{table}

\subsubsection{Performance evaluation}
For each wind farm, we select the last 120 days as our testing period, which corresponds to 120 unique test samples consisting of 24 time steps each. 
The rest of the dataset is used for model training and we use the last 10\% of the training data as our validation set. 
After identifying the parameters that leads to the best forecasting performance using the validation set, we merge the validation set back to the training set, and retrain the models on this merged dataset.
Note that we use the same test sets for global forecasting models as well, therefore, the predictions from individual and global forecasting models are directly comparable.
We employed a two-sided paired t-test for pairwise comparison of forecasting models and understand the statistical significance of the performance improvements attributed to each model.
 

\subsection{Results}
In this section, we provide results from our numerical study, and discuss our findings for the wind power forecasting task.
First, we compare the CNN architectures against other well-known time series forecasting methods.
Then, we examine the impact of incorporating sequential data to the forecasting models, and we assess the effectiveness of Conv2D as feature extractors.

\subsubsection{Comparison of CNNs against other forecasting methods}
Table~\ref{tab:results_summary_without_lags} summarizes the average ND and NRMSE values of the time series forecasting algorithms trained on spatial data alone (i.e., temporal information is excluded). 
The global forecasting approach (i.e., training a model using the combined wind farm dataset) benefits the CNN model the most, reducing average ND values from 0.281 to 0.268.
We note that, when trained individually for each wind farm, CNN performance is worse than other models such as LGBM and ET.
This could be attributed to the fact that complex CNN models often require large training samples to fully capture the nonlinearities within the data \citep{alzubaidi2021review}, and when trained individually for each farm, there might not be enough data instances to generalize the learning.
On the other hand, baseline forecasting models either do not benefit from the global forecasting approach or experience a slight decline in average ND and NRMSE values.
That is, these models are not able to make use of \textit{transfer learning} as effectively and they are not able to combine information coming in from multiple farms to extract meaningful information in the combined dataset.
In terms of average ND and NRMSE values, LGBM and ET perform similarly.
We observe that out of the seven wind farms, CNN is able to outperform other models on farms 1, 3, 5 and 7 in terms of average ND values.
This result indicates that each wind farm exhibits different behaviour and possesses different data characteristics, implying that no single model is best suited for universal wind power forecasting for our dataset. 
\setlength{\tabcolsep}{7.5pt} 
\renewcommand{\arraystretch}{1.5}
\begin{table}[!ht]
\centering
\caption{Performance comparison of models trained on spatial data}
\label{tab:results_summary_without_lags}
\resizebox{0.87\textwidth}{!}{
\begin{tabular}{cccccccccc}
 & \multicolumn{4}{c}{Individual} & \multicolumn{1}{l}{} & \multicolumn{4}{c}{Global} \\ \hline
 & CNN & LGBM & ET & LR & \multicolumn{1}{l}{} & CNN & LGBM & ET & LR \\ \hline
\multicolumn{1}{l|}{\textit{(a) ND}} & \multicolumn{1}{l}{} & \multicolumn{1}{l}{} & \multicolumn{1}{l}{} & \multicolumn{1}{l}{} & \multicolumn{1}{l}{} & \multicolumn{1}{l}{} & \multicolumn{1}{l}{} & \multicolumn{1}{l}{} & \multicolumn{1}{l}{} \\
\multicolumn{1}{c|}{WF1} & 0.285 & 0.276 & 0.279 & 0.358 &  & \textbf{0.266} & 0.279 & 0.273 & 0.356 \\
\multicolumn{1}{c|}{WF2} & 0.311 & 0.282 & \textbf{0.279} & 0.313 &  & 0.289 & 0.291 & 0.284 & 0.326 \\
\multicolumn{1}{c|}{WF3} & 0.289 & 0.275 & 0.275 & 0.326 &  & \textbf{0.271} & 0.285 & 0.291 & 0.322 \\
\multicolumn{1}{c|}{WF4} & 0.226 & 0.204 & 0.201 & 0.253 &  & 0.197 & 0.213 & \textbf{0.195} & 0.269 \\
\multicolumn{1}{c|}{WF5} & 0.259 & 0.261 & 0.261 & 0.298 &  & \textbf{0.248}\textsuperscript{\textdagger} & 0.264 & 0.268 & 0.289 \\
\multicolumn{1}{c|}{WF6} & 0.291 & 0.298 & 0.295 & 0.348 &  & 0.304 & 0.293 & \textbf{0.286} & 0.348 \\
\multicolumn{1}{c|}{WF7} & 0.305 & 0.309 & 0.312 & 0.372 &  & \textbf{0.300}\textsuperscript{\textdagger} & 0.309 & 0.323 & 0.411 \\ \hline
\multicolumn{1}{c|}{Average} & 0.281 & 0.272 & 0.272 & 0.324 &  & \textbf{0.268} & 0.276 & 0.274 & 0.332 \\ \hline
\multicolumn{1}{l|}{\textit{(b) NRMSE}} & \multicolumn{1}{l}{} & \multicolumn{1}{l}{} & \multicolumn{1}{l}{} & \multicolumn{1}{l}{} & \multicolumn{1}{l}{} & \multicolumn{1}{l}{} & \multicolumn{1}{l}{} & \multicolumn{1}{l}{} & \multicolumn{1}{l}{} \\
\multicolumn{1}{c|}{WF1} & 0.361 & 0.352 & 0.354 & 0.431 &  & \textbf{0.339} & 0.361 & 0.349 & 0.439 \\
\multicolumn{1}{c|}{WF2} & 0.394 & 0.354 & \textbf{0.347}\textsuperscript{\textdagger} & 0.375 &  & 0.363 & 0.363 & 0.353 & 0.389 \\
\multicolumn{1}{c|}{WF3} & 0.368 & 0.342 & \textbf{0.339} & 0.385 &  & 0.341 & 0.353 & 0.355 & 0.380 \\
\multicolumn{1}{c|}{WF4} & 0.291 & 0.260 & 0.251 & 0.307 &  & 0.256 & 0.266 & \textbf{0.247} & 0.325 \\
\multicolumn{1}{c|}{WF5} & 0.336 & 0.328 & 0.326 & 0.359 &  & \textbf{0.313}\textsuperscript{\textdagger} & 0.332 & 0.331 & 0.352 \\
\multicolumn{1}{c|}{WF6} & 0.363 & 0.365 & 0.359 & 0.413 &  & 0.373 & 0.359 & \textbf{0.349} & 0.414 \\
\multicolumn{1}{c|}{WF7} & 0.387 & 0.385 & 0.381 & 0.441 &  & \textbf{0.368} & 0.383 & 0.389 & 0.486 \\ \hline
\multicolumn{1}{c|}{Average} & 0.357 & 0.341 & 0.337 & 0.387 &  & \textbf{0.336} & 0.345 & 0.339 & 0.398 \\ \hline
\bottomrule
\multicolumn{9}{l}{\quad \ \textsuperscript{\textdagger}: Significant at the 95\% level (2-sided paired t-test)}\\
\end{tabular}
}
\end{table}

\begin{figure}[!ht]
    \centering
    \subfloat[ND] {{\includegraphics[width=0.50\textwidth]{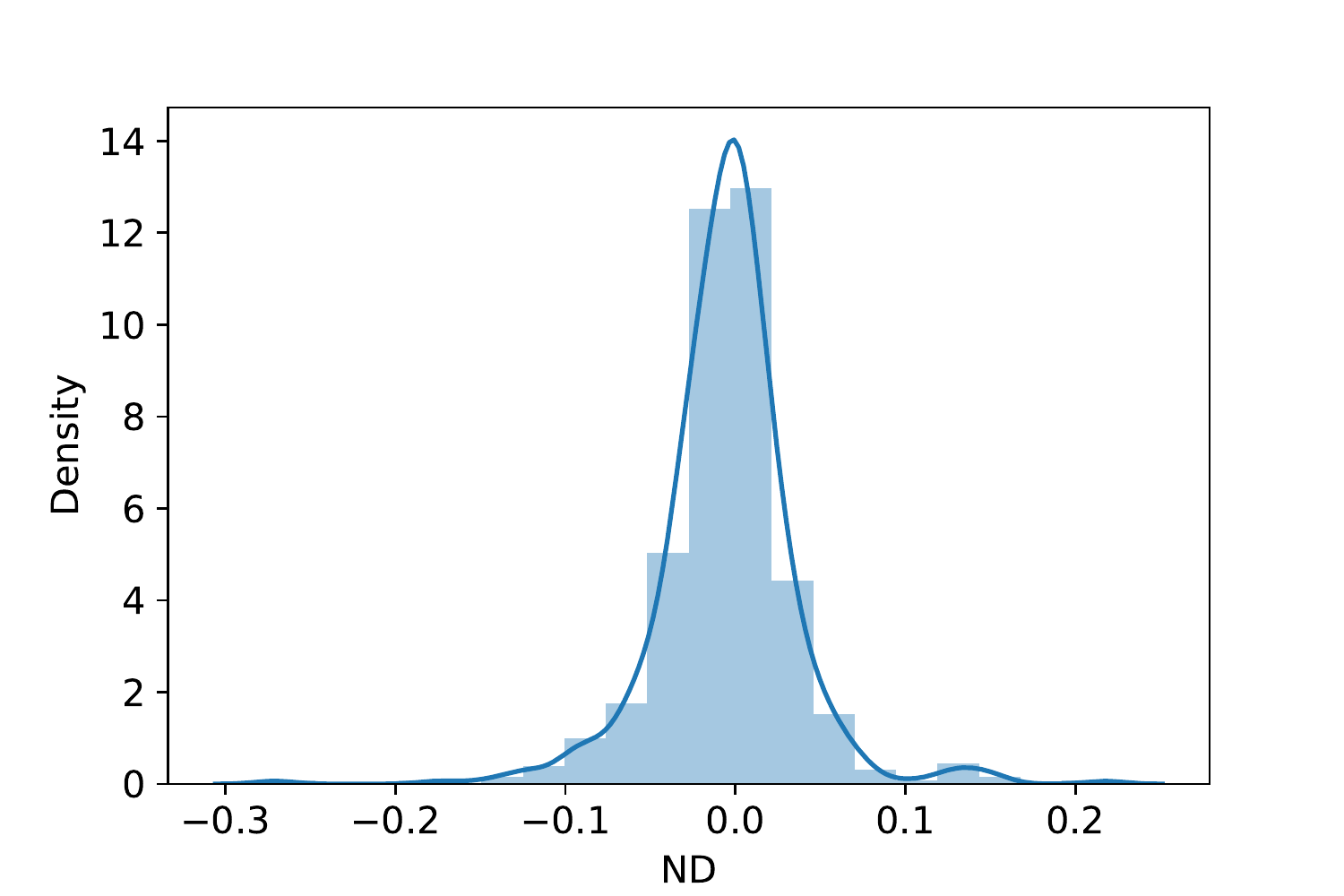}}}
     \subfloat[NRMSE] {{\includegraphics[width=0.5\textwidth]{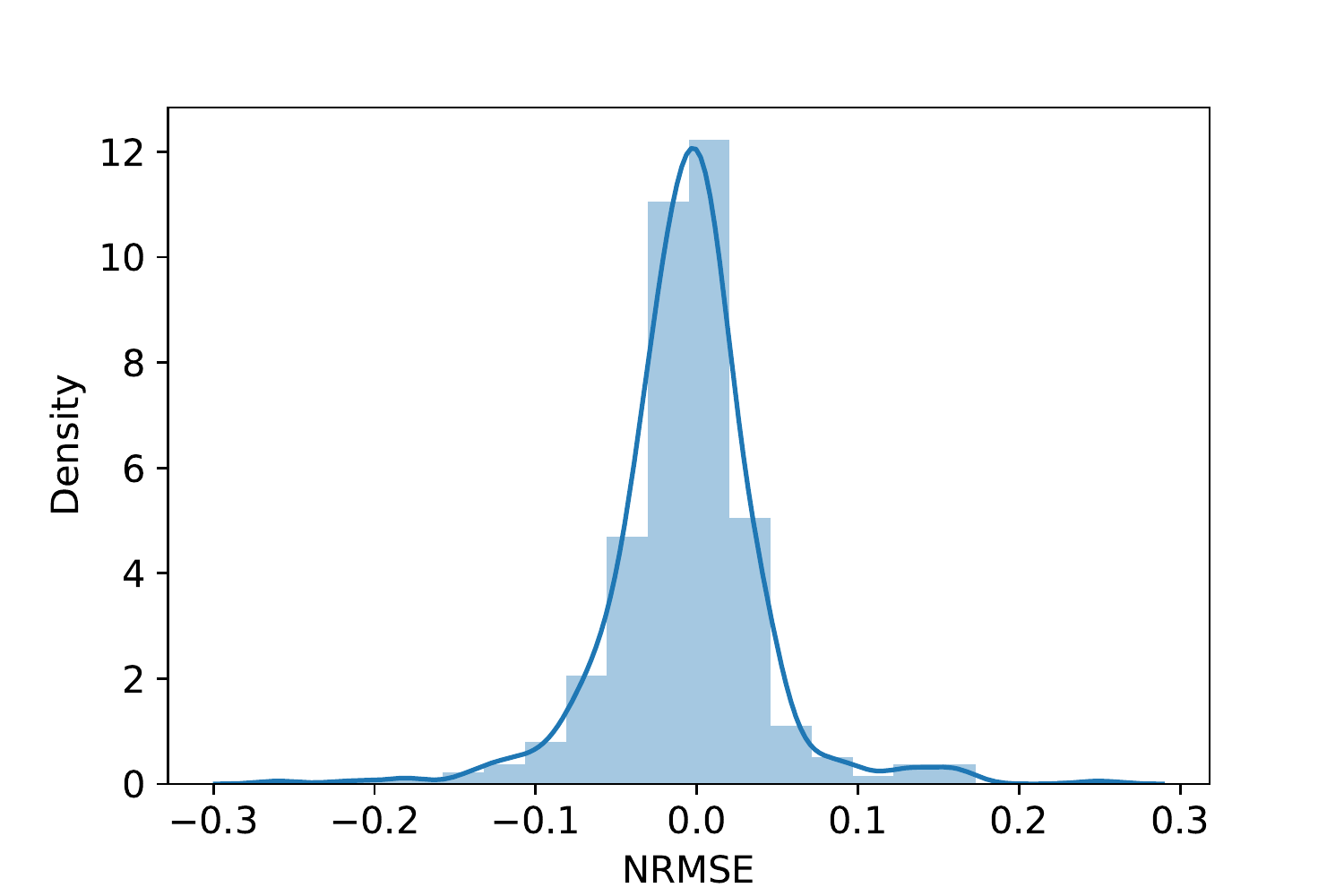}}}
    \caption{Plots showing normal distributions of ND and NRMSE errors making them suitable for two-sided paired t-test}
    \label{fig:conv_structure_visuals}
\end{figure}

To determine whether the performance difference between these models is statistically significant, we conduct a two-sided paired t-test and adopt a similar approach to \citet{oreshkin2019n}. 
That is, we employ a statistical procedure to identify whether the mean difference between two sets of results that are compared against each other is zero. 
Figure~\ref{fig:conv_structure_visuals} shows the distributions of ND and NRMSE errors, indicating that these results are suitable for conducting two-sided paired t-test.
We compare the performance of globally trained CNN against ET model trained over individual wind farms, as these two models show a similar level of performance. 
The null ($H_{0}$) and alternative ($H_{1}$) hypotheses are characterized as follows:
\begin{itemize}
\centering
    \item $H_{0}$: $\mu^{ET}$ = $\mu^{CNN}$
    \item $H_{1}$: $\mu^{ET}$ $\neq$ $\mu^{CNN}$
\end{itemize}
Here $H_{0}$ signifies that the mean ET and CNN scores are equal, while $H_{1}$ signifies that the mean ET and CNN scores are not equal.
The $p$-values for ND are smaller than 0.05 for wind farms 2 and 5, at 0.038 and 0.037, respectively, and for NRMSE, for wind farms 5 and 7, at 0.011 and 0.045, respectively. 
Therefore only these results can be considered statistically significant.

\subsubsection{Impact of incorporating sequential data}
Table~\ref{tab:results_summary_with_lags} summarizes performance of models when trained on spatial data and lag values of wind power output. 
We observe that while the performance of all the models improve with the inclusion of temporal information, the most significant improvement is witnessed for the combined CNN-RNN architecture. 
This could be due to LSTM's ability to preserve long sequential information of the lag inputs, which is not possible for the baseline forecasting models such as LGBM and ET.
We note that CNN-RNN improves the performance over the CNN model, with average ND values improving from 0.268 to 0.249 for the global forecasting case.

\setlength{\tabcolsep}{7.5pt} 
\renewcommand{\arraystretch}{1.5}
\begin{table}[!ht]
\centering
\caption{Performance comparison of models trained on spatial and temporal data}
\label{tab:results_summary_with_lags}
\begingroup
\resizebox{0.95\textwidth}{!}{
\begin{tabular}{cccccccccc}
 & \multicolumn{4}{c}{Individual} & \multicolumn{1}{l}{} & \multicolumn{4}{c}{Global} \\ \hline
 & CNN-RNN & LGBM & ET & LR & \multicolumn{1}{l}{} & CNN-RNN & LGBM & ET & LR \\ \hline
\multicolumn{1}{l|}{\textit{(a) ND}} & \multicolumn{1}{l}{} & \multicolumn{1}{l}{} & \multicolumn{1}{l}{} & \multicolumn{1}{l}{} & \multicolumn{1}{l}{} & \multicolumn{1}{l}{} & \multicolumn{1}{l}{} & \multicolumn{1}{l}{} & \multicolumn{1}{l}{} \\
\multicolumn{1}{c|}{WF1} & 0.262 & 0.266 & 0.265 & 0.324 &  & \textbf{0.256} & 0.260 & 0.261 & 0.329 \\
\multicolumn{1}{c|}{WF2} & 0.285 & 0.275 & \textbf{0.271} & 0.298 &  & 0.282 & 0.275 & 0.278 & 0.305 \\
\multicolumn{1}{c|}{WF3} & 0.269 & 0.266 & 0.270 & 0.308 &  & \textbf{0.259} & 0.274 & 0.283 & 0.306 \\
\multicolumn{1}{c|}{WF4} & 0.197 & \textbf{0.185}\textsuperscript{\textdagger} & 0.190 & 0.235 &  & 0.188 & 0.207 & 0.195 & 0.252 \\
\multicolumn{1}{c|}{WF5} & 0.257 & 0.240 & 0.248 & 0.276 &  & \textbf{0.237}\textsuperscript{\textdagger} & 0.260 & 0.256 & 0.269 \\
\multicolumn{1}{c|}{WF6} & 0.270 & 0.283 & 0.279 & 0.322 &  & \textbf{0.258} & 0.275 & 0.271 & 0.322 \\
\multicolumn{1}{c|}{WF7} & 0.271 & 0.290 & 0.300 & 0.334 &  & \textbf{0.263}\textsuperscript{\textdagger} & 0.290 & 0.306 & 0.352 \\ \hline
\multicolumn{1}{c|}{Average} & 0.259 & 0.258 & 0.260 & 0.300 &  & \textbf{0.249}\textsuperscript{\textdagger} & 0.263 & 0.264 & 0.305 \\ \hline
\multicolumn{1}{l|}{\textit{(b) NRMSE}} & \multicolumn{1}{l}{} & \multicolumn{1}{l}{} & \multicolumn{1}{l}{} & \multicolumn{1}{l}{} & \multicolumn{1}{l}{} & \multicolumn{1}{l}{} & \multicolumn{1}{l}{} & \multicolumn{1}{l}{} & \multicolumn{1}{l}{} \\
\multicolumn{1}{c|}{WF1} & 0.332 & 0.337 & 0.337 & 0.390 &  & \textbf{0.331} & 0.332 & 0.331 & 0.407 \\
\multicolumn{1}{c|}{WF2} & 0.354 & 0.344 & \textbf{0.336} & 0.360 &  & 0.354 & 0.345 & 0.346 & 0.368 \\
\multicolumn{1}{c|}{WF3} & 0.334 & 0.333 & \textbf{0.330} & 0.367 &  & \textbf{0.330} & 0.342 & 0.347 & 0.365 \\
\multicolumn{1}{c|}{WF4} & 0.247 & 0.238 & \textbf{0.238} & 0.288 &  & 0.242 & 0.260 & 0.246 & 0.306 \\
\multicolumn{1}{c|}{WF5} & 0.335 & 0.306 & 0.312 & 0.337 &  & \textbf{0.304} & 0.325 & 0.319 & 0.334 \\
\multicolumn{1}{c|}{WF6} & 0.332 & 0.346 & 0.340 & 0.384 &  & \textbf{0.324} & 0.338 & 0.330 & 0.386 \\
\multicolumn{1}{c|}{WF7} & 0.341 & 0.360 & 0.362 & 0.405 &  & \textbf{0.324}\textsuperscript{\textdagger} & 0.357 & 0.369 & 0.424 \\ \hline
\multicolumn{1}{c|}{Average} & 0.325 & 0.323 & 0.322 & 0.362 &  & \textbf{0.316}\textsuperscript{\textdagger} & 0.328 & 0.327 & 0.370 \\ \hline
\bottomrule
\multicolumn{9}{l}{\quad \ \textsuperscript{\textdagger}: Significant at the 95\% level (2-sided paired t-test)}\\
\end{tabular}
}
\endgroup
\end{table}

We consider the following hypothesis tests to assess the statistical significance of the performance differences between CNN-RNN and ET models:
\begin{itemize}
\centering
    \item $H_{0}$: $\mu^{ET}$ = $\mu^{\textit{CNN-RNN}}$
    \item $H_{1}$: $\mu^{ET}$ $\neq$ $\mu^{\textit{CNN-RNN}}$
\end{itemize}
where $H_{0}$ signifies that the mean performance values for ET and CNN-RNN scores are equal, while $H_{1}$ characterizes the alternative hypothesis. 
The CNN-RNN architecture is able to outperform other methods overall at 95\% statistical significance level, with $p$-values of 0.001 for ND and 0.047 for NRMSE. 
For the ND performance over individual wind farms, farms 3, 5 and 7 indicate statistically significant improvements for CNN-RNN with $p$-values of 0.024, 0.047 and 0.002, respectively, while, for NRMSE, only the performance for farm 7 is statistically significant with $p$-value of 0.003. 

\begin{figure*}[!ht]
    \begin{center}
    \subfloat[Test sample 1] {\includegraphics[width=0.4950\textwidth]{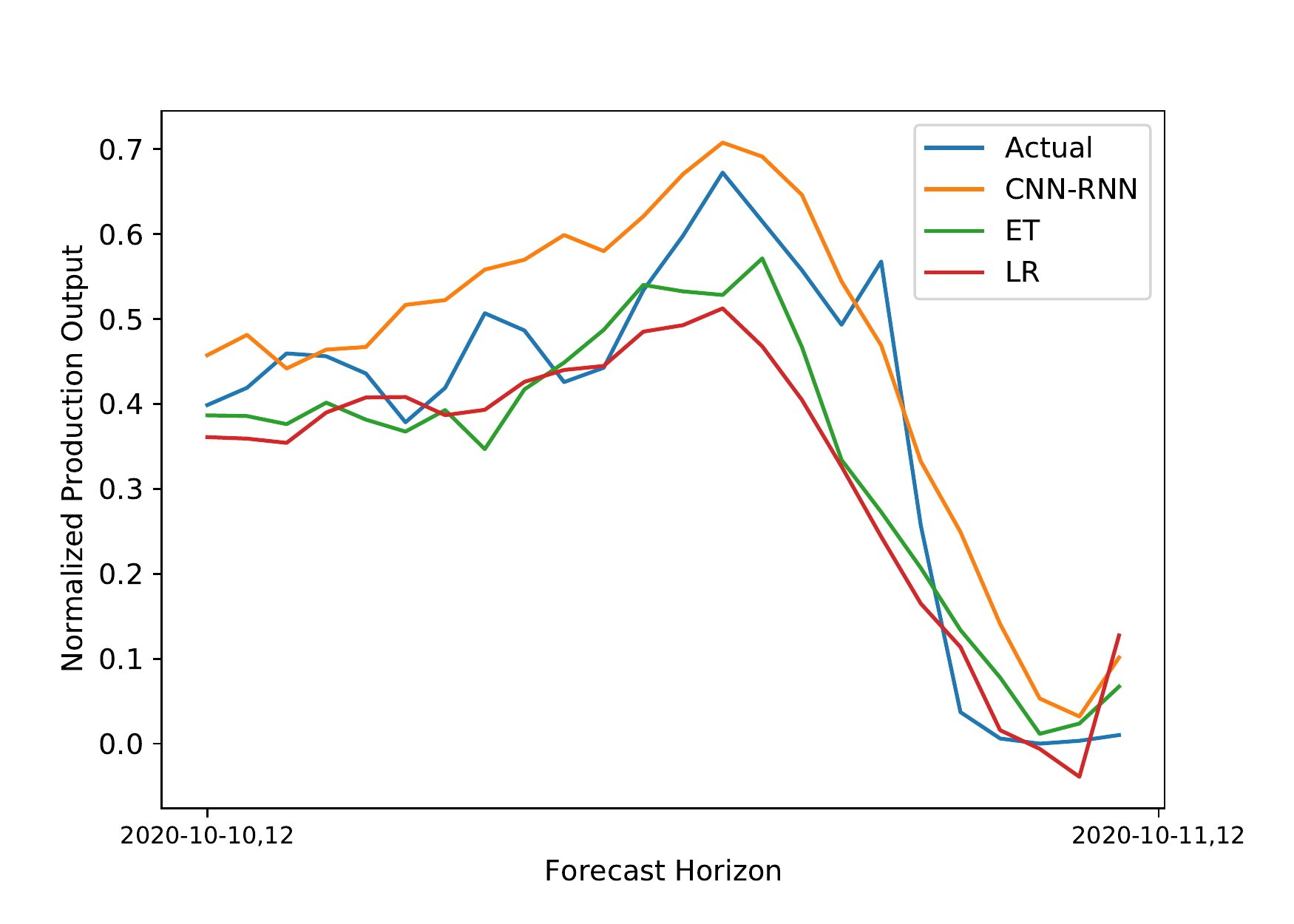}}
    \subfloat[Test sample 2] {\includegraphics[width=0.4950\textwidth]{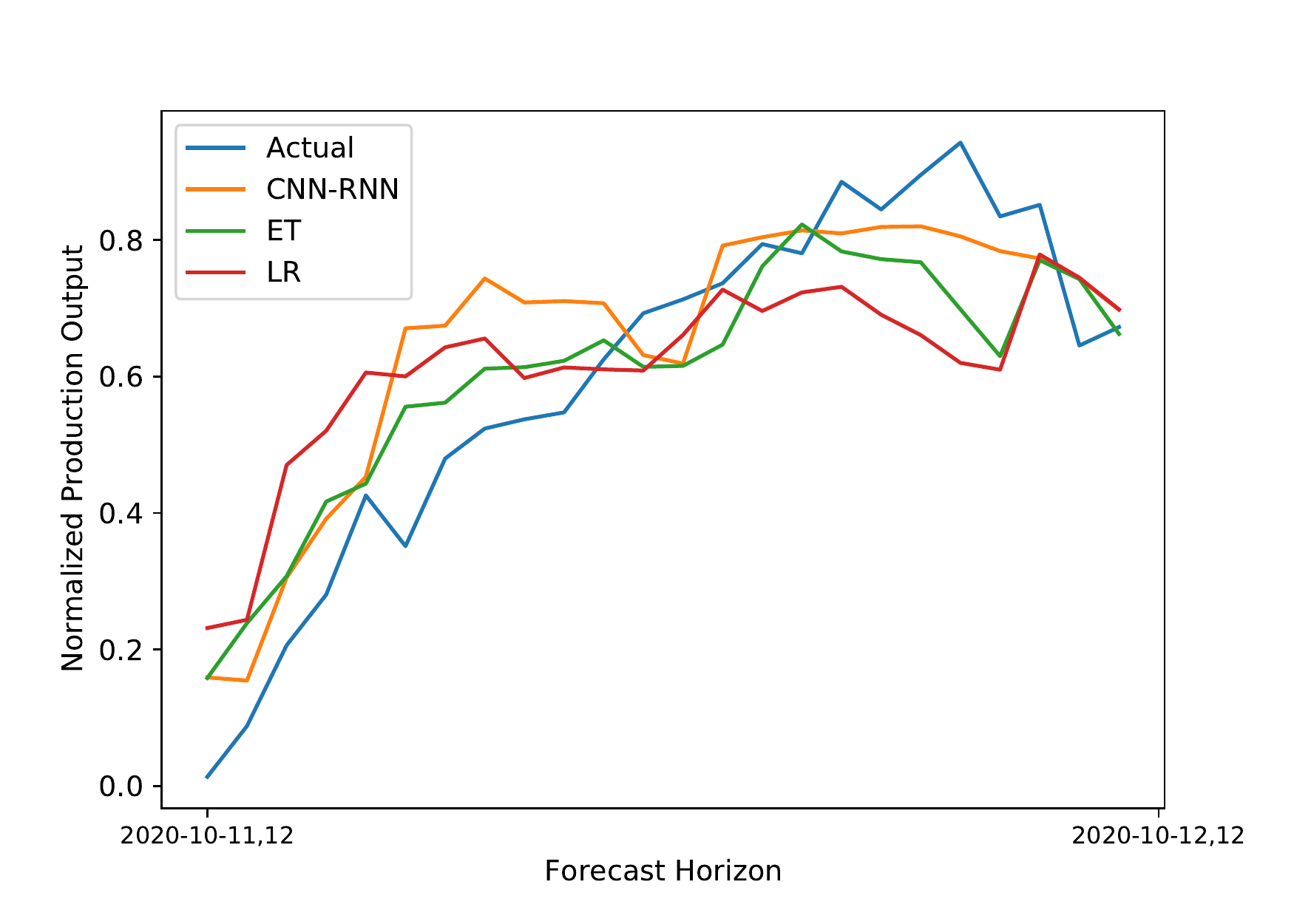}}
    \\
     \subfloat[Test sample 3] {\includegraphics[width=0.4950\textwidth]{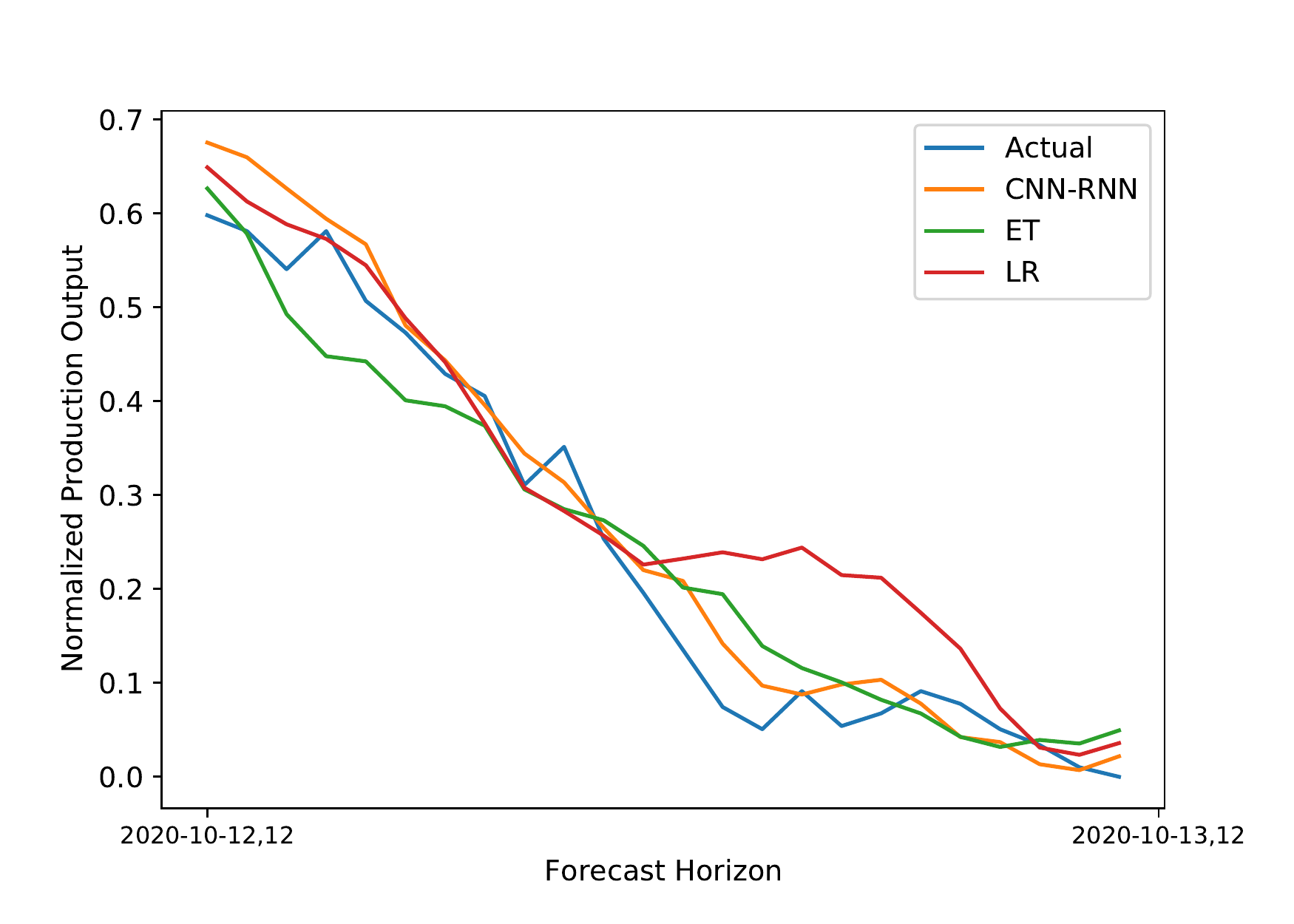}}
    \subfloat[Test sample 4] {\includegraphics[width=0.4950\textwidth]{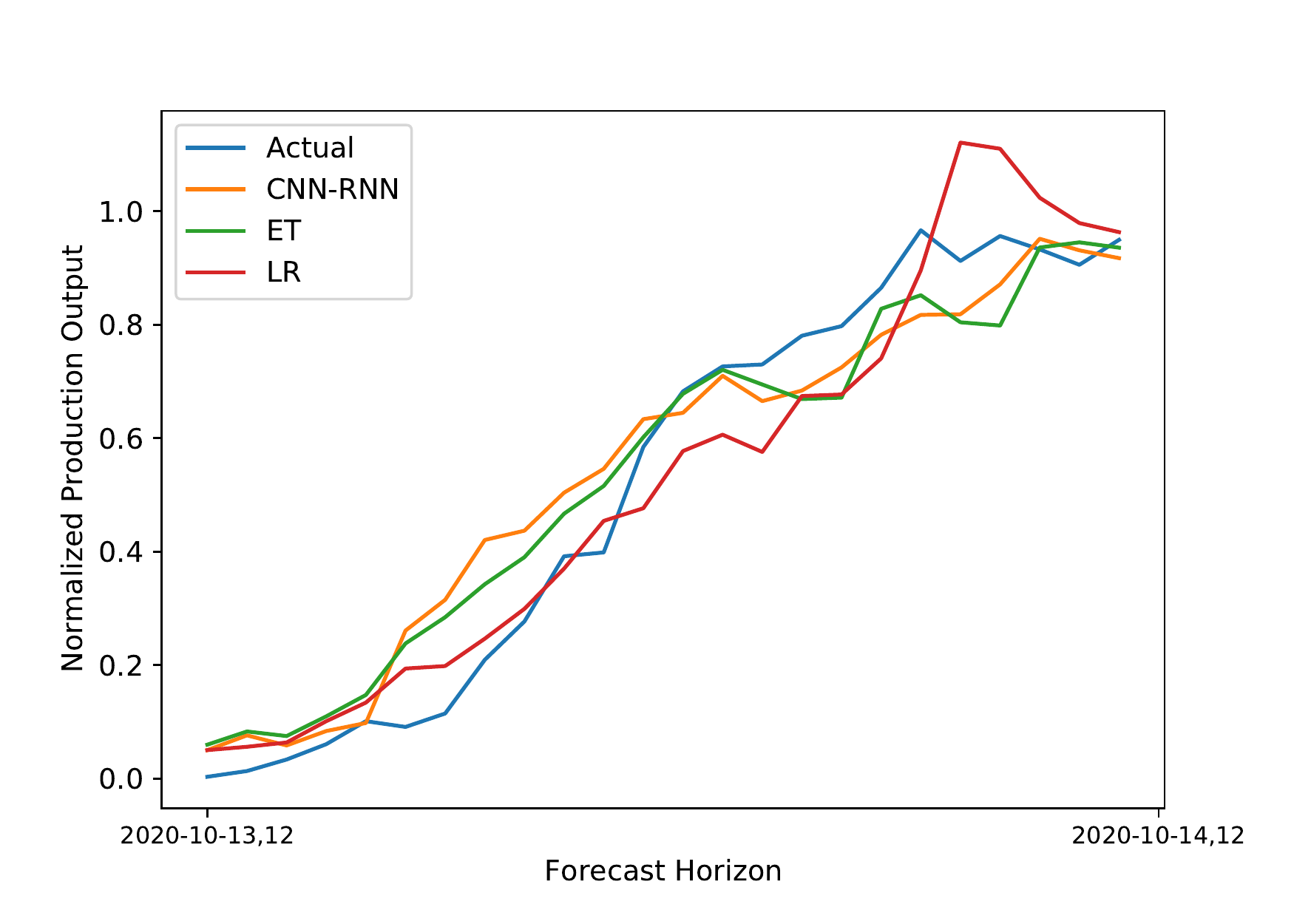}}
    \\
     \subfloat[Test sample 5] {\includegraphics[width=0.4950\textwidth]{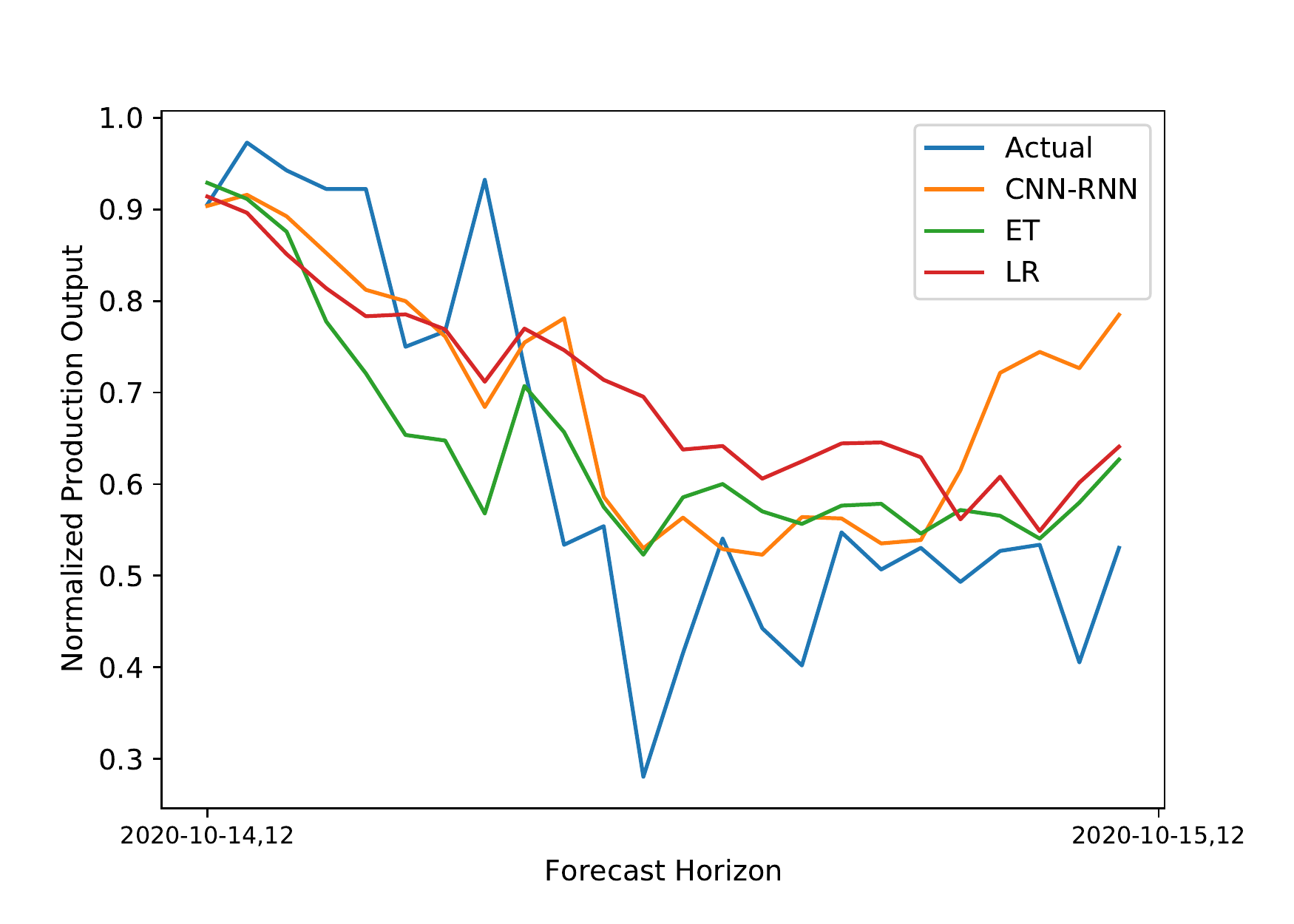}}
    \subfloat[Test sample 6] {\includegraphics[width=0.4950\textwidth]{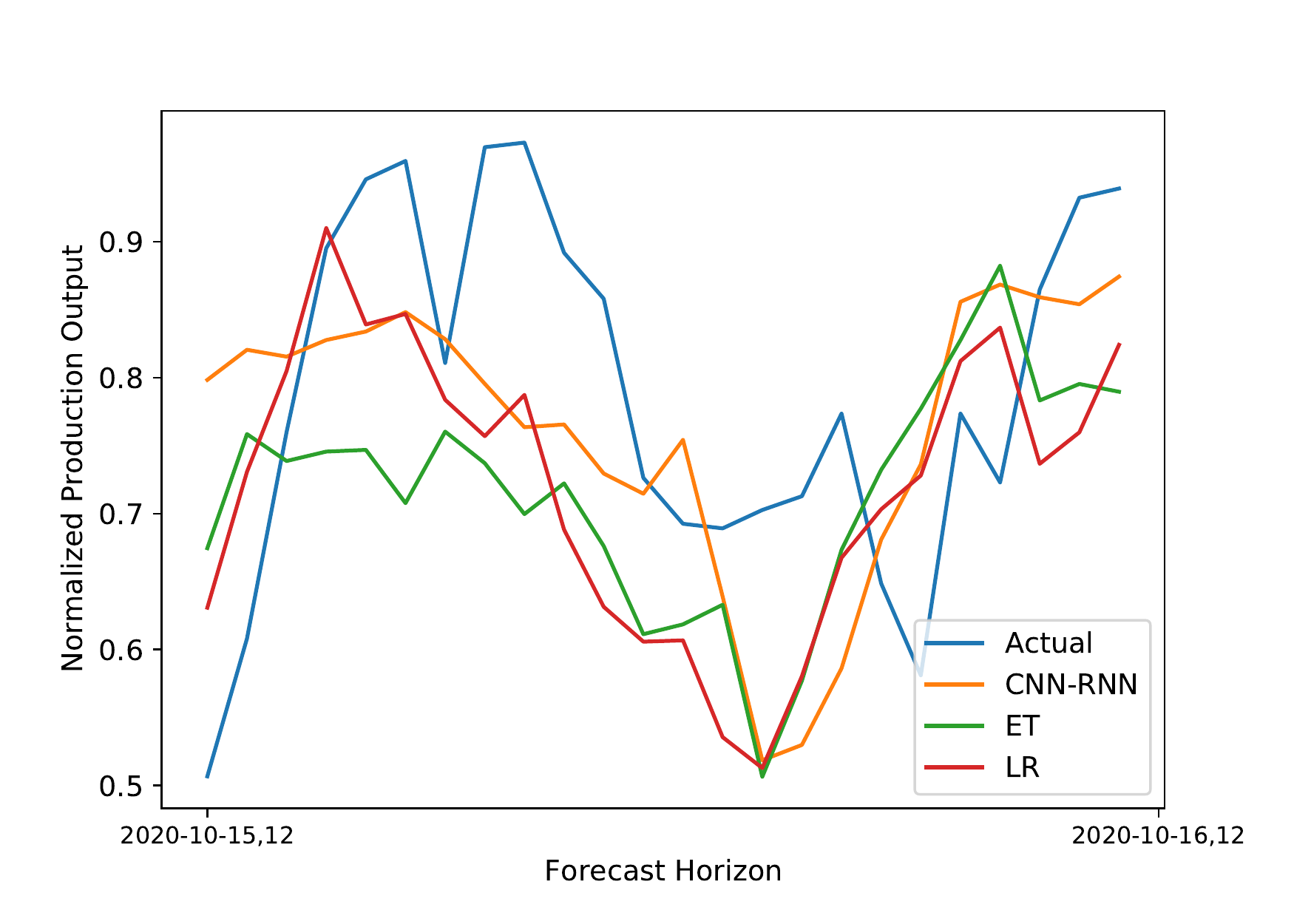}}
 \caption{Visual comparison of model predictions across six 24-hr long test samples }
    \label{fig:sample_predictions}
    \end{center}
\end{figure*}

Figure~\ref{fig:sample_predictions} demonstrates model predictions across six unique sample test batches. 
We observe that while none of the models provide highly conformal predictions, the forecasts largely follow the trends for the ground truth (Actual) values.

\begin{figure}[!ht]
    \centering
    \subfloat[ND] {{\includegraphics[width=0.9\textwidth]{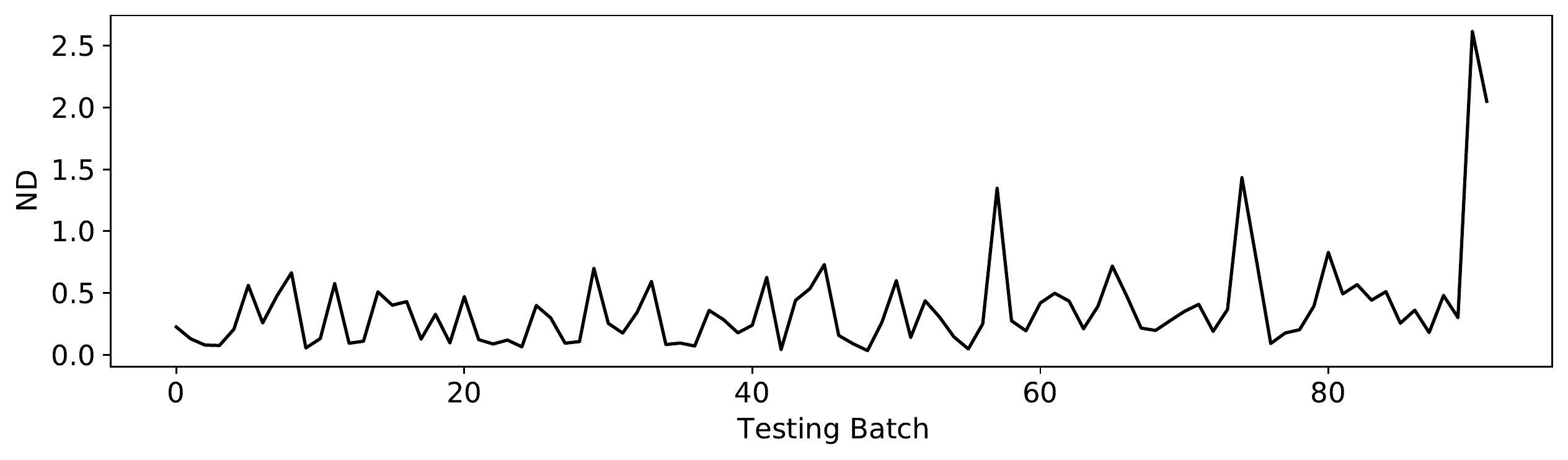}}}\hspace{0.11in}
     \subfloat[NRMSE] {{\includegraphics[width=0.9\textwidth]{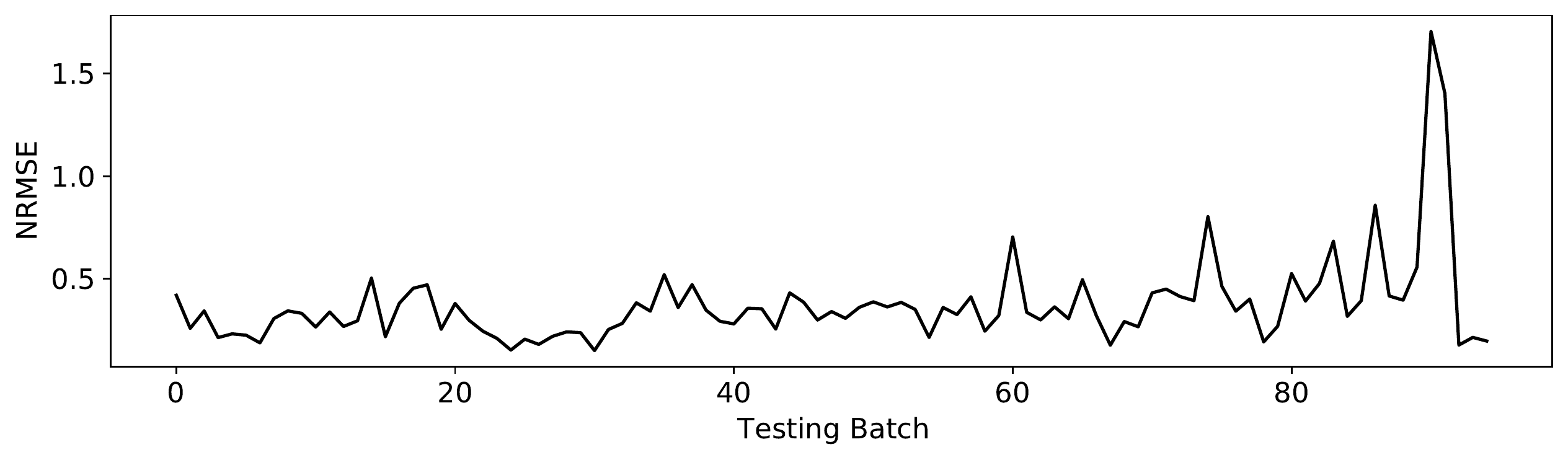} \label{fig:totalnd_nrmse}}}
    \caption{Average ND and NRMSE across all test batches for all the models}
    \label{fig:detailed_errors}
\end{figure}

Figure~\ref{fig:detailed_errors} illustrates the average ND and NRMSE errors over all the trained forecasting models across the test batches. 
That is, errors from all the different models are averaged for each test batch, with the goal of understanding which test batches are more difficult to predict, and whether there are any visible outliers. 
We find that while the ND error is below 1.0, and NRMSE is below 0.5 for the first 50 test batch samples, the errors drastically increase towards the last set of test batches. 
This shift in performance is typically expected for time series forecasting tasks, as the further the test predictions are from the last training batch set, the more difficult it is to obtain accurate results.
Figure~\ref{fig:detailed_errors} also shows that the performance across the different batches inherits high variance, with the spikes indicating high noise in the testing performance. 
This observation indicates that our datasets are highly complicated and they contain a significant amount of noise. 

\subsubsection{Using Conv2D as spatial feature extractor}
We next examine whether spatial features extracted from Conv2D layers benefit the performance of LGBM.
Specifically, when added to LGBM, we expect the extracted features to enhance the prediction performance because the original features are kept in the LGBM training. 
On the other hand, replacing the fully connected layers of a CNN with LGBM help might achieve better results than a traditional CNN. 
For this experiment, we use the global forecasting approach and convolutional 2D layers, while LGBM is trained for each wind farm individually. Table~\ref{tab:conv2d_lgbm_impr} shows the absolute change in performance for Conv2D + LGBM versus LGBM and CNN. 
The downward and upward arrows represent a drop or increase in error, respectively. 

\setlength{\tabcolsep}{7.5pt} 
\renewcommand{\arraystretch}{1.37} 
\begin{table}[!ht]
\centering
\caption{Performance improvements for Conv2D + LGBM}
\label{tab:conv2d_lgbm_impr}
\subfloat[ND \label{tab:conv2d_nd}]{
\resizebox{0.979\textwidth}{!}{
\begin{tabular}{c|ccccccc|r|}
\cline{2-9}
 & \textbf{WF1} & \textbf{WF2} & \textbf{WF3} & \textbf{WF4} & \textbf{WF5} & \textbf{WF6} & \textbf{WF7} & \multicolumn{1}{c|}{\textbf{Average}} \\ \hline
\multicolumn{1}{|c|}{Conv2D + LGBM} & 0.270 & 0.283 & 0.276 & 0.199 & 0.260 & 0.286 & 0.299 & 0.267 \\
\multicolumn{1}{|c|}{vs LGBM} & \multicolumn{1}{r}{\textbf{$\downarrow$ 0.006}} & \multicolumn{1}{r}{$\uparrow$ 0.001} & \multicolumn{1}{r}{$\uparrow$ 0.001} & \multicolumn{1}{r}{\textbf{$\downarrow$ 0.005}} & \multicolumn{1}{r}{\textbf{$\downarrow$ 0.001}} & \multicolumn{1}{r}{\textbf{$\downarrow$ 0.012}\textsuperscript{\textdagger}} & \multicolumn{1}{r|}{\textbf{$\downarrow$ 0.010}\textsuperscript{\textdagger}} & $\downarrow$ \textbf{0.004}\textsuperscript{\textdagger} \\
\multicolumn{1}{|c|}{vs CNN} & \multicolumn{1}{r}{$\uparrow$ 0.004} & \multicolumn{1}{r}{\textbf{$\downarrow$ 0.006}} & \multicolumn{1}{r}{$\uparrow$ 0.005} & \multicolumn{1}{r}{$\uparrow$ 0.002} & \multicolumn{1}{r}{$\uparrow$ 0.012\textsuperscript{\textdagger}} & \multicolumn{1}{r}{\textbf{$\downarrow$ 0.018\textsuperscript{\textdagger}}} & \multicolumn{1}{r|}{\textbf{$\downarrow$ 0.001}} & \textbf{$\downarrow$ 0.001} \\ \hline

\bottomrule
\multicolumn{9}{l}{\quad \ \textsuperscript{\textdagger}: Significant at the 95\% level (2-sided paired t-test)}\\
\end{tabular}
}
}\\
\subfloat[NRMSE \label{tab:conv2d_nrmse}]{
\resizebox{0.979\textwidth}{!}{
\begin{tabular}{c|ccccccc|r|}
\cline{2-9}
 & \textbf{WF1} & \textbf{WF2} & \textbf{WF3} & \textbf{WF4} & \textbf{WF5} & \textbf{WF6} & \textbf{WF7} & \multicolumn{1}{c|}{\textbf{Average}} \\ \hline
\multicolumn{1}{|c|}{Conv2D + LGBM} & 0.343 & 0.355 & 0.343 & 0.255 & 0.328 & 0.352 & 0.371 & 0.335 \\
\multicolumn{1}{|c|}{vs LGBM} & \multicolumn{1}{r}{\textbf{$\downarrow$ 0.009}} & \multicolumn{1}{r}{$\uparrow$ 0.001} & \multicolumn{1}{r}{$\uparrow$ 0.001} & \multicolumn{1}{r}{\textbf{$\downarrow$ 0.005}} & \multicolumn{1}{r}{0.000} & \multicolumn{1}{r}{\textbf{$\downarrow$ 0.013\textsuperscript{\textdagger}}} & \multicolumn{1}{r|}{$\downarrow$ 0.014\textsuperscript{\textdagger}} & $\downarrow$ \textbf{0.006}\textsuperscript{\textdagger}\\
\multicolumn{1}{|c|}{vs CNN} & \multicolumn{1}{r}{$\uparrow$ 0.004} & \multicolumn{1}{r}{\textbf{$\downarrow$ 0.013}} & \multicolumn{1}{r}{$\uparrow$ 0.002} & \multicolumn{1}{r}{\textbf{$\downarrow$ 0.001}} & \multicolumn{1}{r}{$\uparrow$ 0.015\textsuperscript{\textdagger}} & \multicolumn{1}{r}{\textbf{$\downarrow$ 0.021\textsuperscript{\textdagger}}} & \multicolumn{1}{r|}{$\uparrow$ 0.003} & \textbf{$\downarrow$ 0.002} \\ \hline

\bottomrule
\multicolumn{9}{l}{\quad \ \textsuperscript{\textdagger}: Significant at the 95\% level (2-sided paired t-test)}\\
\end{tabular}
}
}
\end{table}

We conduct two hypothesis tests to measure the significance of performance improvements attributed to using Conv2D as spatial feature extractor.
The null and alternative hypothesis for the first hypothesis test is as follows:
\begin{itemize}
\centering
    \item $H_{0}$: $\mu^{LGBM}$ = $\mu^{\textit{Conv2D + LGBM}}$
    \item $H_{1}$: $\mu^{LGBM}$ $\neq$ $\mu^{\textit{Conv2D + LGBM}}$
\end{itemize}
where $H_{0}$ signifies that the mean LGBM and Conv2D + LGBM performance values are equal, while $H_{1}$ corresponds to the alternative hypothesis.
We observe that Conv2D + LGBM reduces average ND error by 0.004, and average NRMSE error by 0.006 when compared against LGBM.
We find that the $p$-values for this test for ND and NRMSE are 0.004 and 0.002, respectively, indicating that the performance improvements are statistically significant.
For individual wind farms, Conv2D + LGBM is able to outperform LGBM on wind farms 6 and 7 with $p$-values of 0.045 and 0.008, respectively, for ND and 0.043 and 0.003, respectively, for NRMSE. 

The null and alternative hypothesis for the second hypothesis test is as follows:
\begin{itemize}
\centering
    \item $H_{0}$: $\mu^{CNN}$ = $\mu^{\textit{Conv2D + LGBM}}$
    \item $H_{1}$: $\mu^{CNN}$ $\neq$ $\mu^{\textit{Conv2D + LGBM}}$
\end{itemize}
where $H_{0}$ signifies that the mean CNN and Conv2D + LGBM scores are equal, while $H_{1}$ corresponds to the alternative hyppothesis. 
We observe smaller improvements for Conv2D + LGBM over CNN (0.001 and 0002, for ND and NRMSE, respectively), and with $p$-values of 0.99 and 0.83 for ND and NRMSE, respectively, the improvements were not found to be significant (i.e., fail to reject the null hypothesis).
We also find statistically significant improvements attributed to Conv2D + LGBM over CNN for wind farms 5 and 6 with $p$-values of 0.023 and 0.018 for ND, and 0.012 and 0.010 for NRMSE, respectively.

\begin{figure}[!ht]
    \centering
    \subfloat[ND] {{\includegraphics[width=0.50\textwidth]{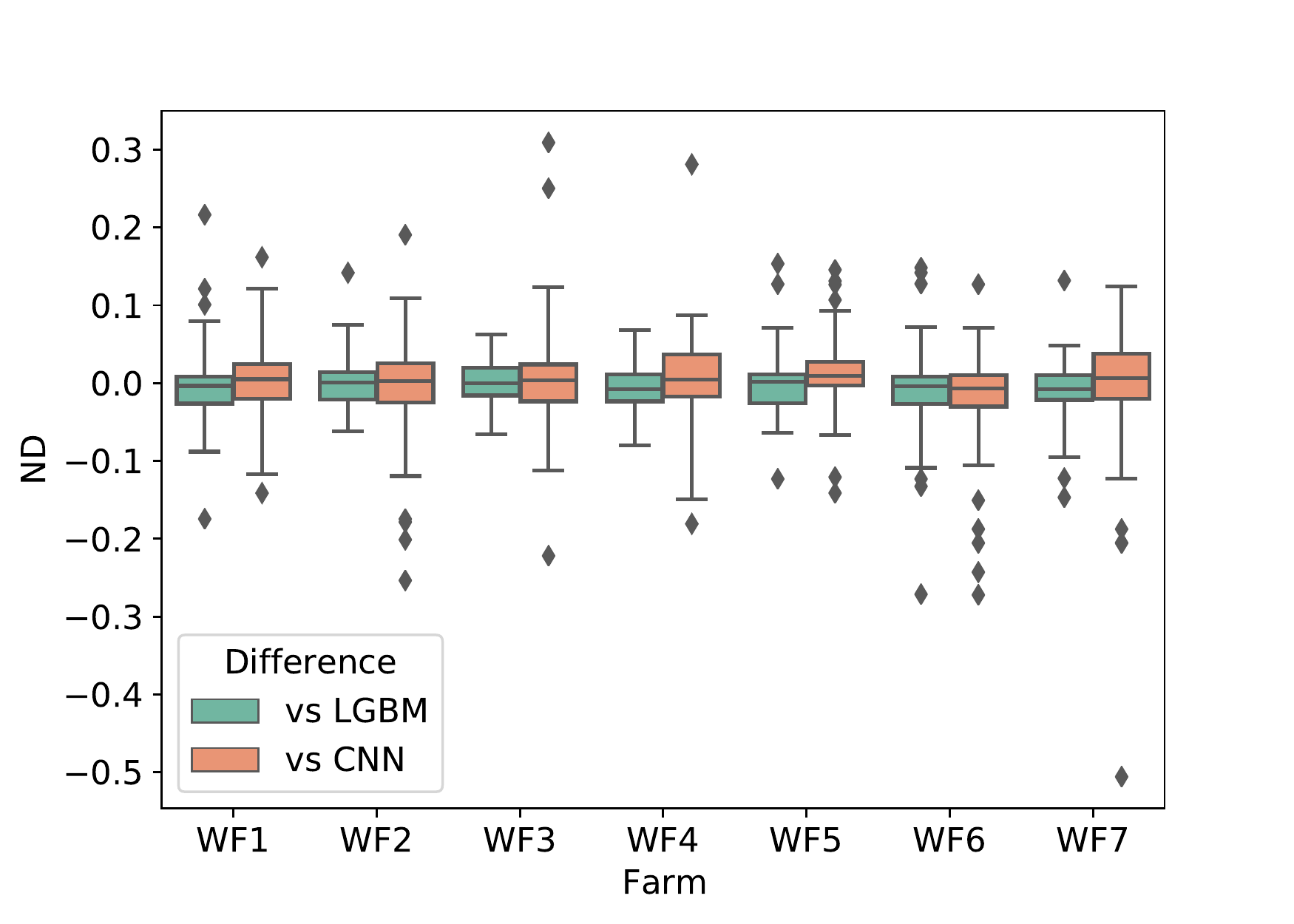}}} 
     \subfloat[NRMSE] {{\includegraphics[width=0.50\textwidth]{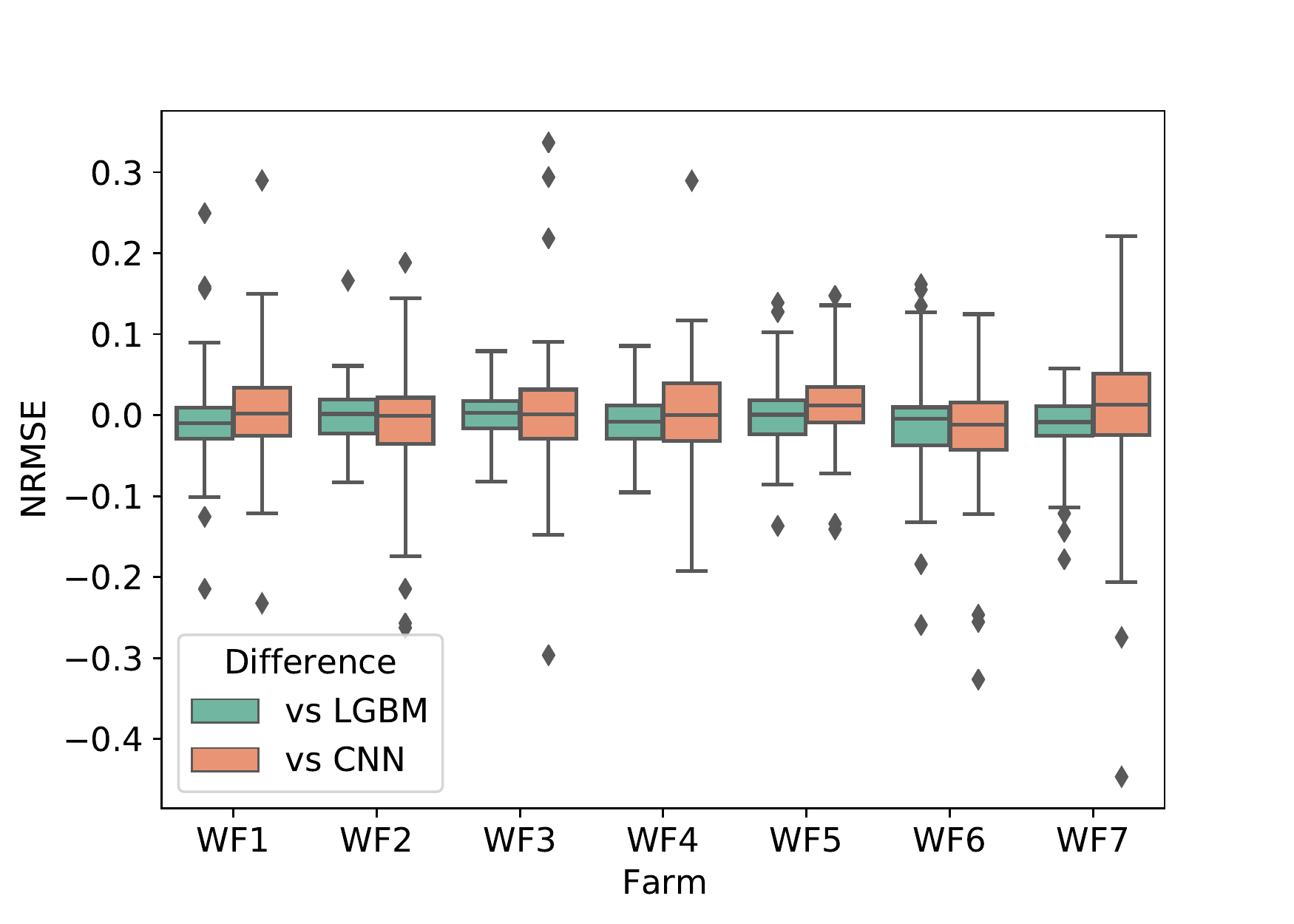} \label{fig:statistical_ress_newb}}}
    \caption{Statistical distribution of performance differences for Conv2D + LGBM vs LGBM and CNN }
    \label{fig:statistical_ress_new}
\end{figure}

These results show that, in most cases, it is possible to utilize convolutional layers to extract spatial information from location-based features and use that additional information to further improve the performance of a strong ensemble-based regressor (e.g., LGBM). 
Additionally, for some cases, by replacing the fully connected dense network of a CNN architecture with a strong ensemble-based regressor, we are able to achieve better performance than CNN baseline.
Figure~\ref{fig:statistical_ress_new} shows the distribution of the difference in ND and NRMSE errors across the seven wind farms, for LGBM and CNN, when compared against Conv2D + LGBM. 

\section{Conclusions and Discussions}\label{sec:conclusions}

Wind power forecasting and ramp event prediction have garnered significant interest as they can be used for various purposes in practice such as taking preventative actions to reduce equipment damage and improving operational efficiency within the power grid. 
This work focuses on a complex combined architecture involving CNNs and RNNs to extract useful information from an ultra-wide input matrix that consists of entries from multiple NWP models, wind farms, geographical locations and atmospheric levels, to make a day-ahead wind power and ramp event predictions. 
Our analysis highlights the capabilities of CNN architectures in extracting spatial information by learning the underlying interdependencies of input data across various locations and RNN's capability towards extracting long-term temporal information. 
We also make use of CNN's spatial feature extraction ability by feeding the extracted features to a tree ensemble-based regressor, namely LGBM, to further boost its performance. 
We conduct numerical studies to assess the impact of global learning on wind power prediction performance.

We evaluate the performance of our models using the ND and NRMSE metrics and employ the two-sided pairwise t-test to assess the significance of performance improvements attributed to the proposed models. 
Our results show that CNN and RNN make effective use of the spatial and temporal information in the dataset and, when combined together to form a complex neural structure, they are able to outperform other ML models including LR, ET and LGBM. 
Our numerical results also show that the Conv2D + LGBM method obtained by feeding the features extracted from CNN to LGBM performs better than standalone usage of LGBM, highlighting the importance of extracting spatial information from location-based features, and in some cases better than CNN, signifying that, for certain datasets, a simpler regression model may better learn from the available data compared to the complex neural network architectures.
We observe that global learning contributes significantly to the performance of CNN and CNN-RNN, as these complex neural networks can leverage increased training set sizes from the combined dataset of multiple wind farms, and they are better able to extract the interdependencies between the related data sources. 

There have been several challenges and limitations to our work. The data sets we incorporated in this study were extremely noisy, with a high variance within the testing batches, making it difficult for us to achieve highly accurate results. 
This issue is very common in medium and long-term wind power forecasting problems, and a significant amount of research is dedicated to improving the forecasting performance with noisy data.
For multi-step ahead ramp classification, we have dealt with a highly imbalanced class distribution, with the ratio of ramp event to no-ramp event being 1:5,673. 
While we adjust the weights of the classes in the ML model training, it is still difficult to achieve satisfactory performance for the ramp detection task. 
Due to the massive scale of our input data combined from multiple sources, it is difficult to perform extensive hyperparameter tuning and experiment with more complex models, e.g., by combining ConvLSTM with our CNN-RNN architecture. 

Several research directions can be considered to extend our work in the future. The proposed CNN-RNN architecture can be further enhanced by adding a ConvLSTM component to it for extracting spatio-temporal information from the historical meteorological features and using TCN along with RNN for temporal feature extraction from the historic wind power outputs. 
This was not possible at present due to the high computational cost involved with this modified architecture. 
Our experiments involving data from seven distinct wind farms show that each wind farm has its own unique characteristics, which may be due to its location or the differences in the wind turbines installed, and that no single model is best suited to make predictions across all the wind farms.
In this regard, an ensemble method can be considered for our forecasting task where final predictions are a combined result of individually and globally trained models with adjusted weights.
As more data becomes available, a continuous latitude and longitude grid can be employed, similar to \citet{kazutoshi2018feature}, to allow for better representation of the spatial feature space.
An augmented out-of-sample technique~\citep{ilic2020augmented} can be employed in order to improve the prediction performance for the test batches that are farthest from the training set.

\bibliographystyle{elsarticle-harv}
\bibliography{main_springer}

\begin{thebibliography}{49}
\expandafter\ifx\csname natexlab\endcsname\relax\def\natexlab#1{#1}\fi
\providecommand{\url}[1]{\texttt{#1}}
\providecommand{\href}[2]{#2}
\providecommand{\path}[1]{#1}
\providecommand{\DOIprefix}{doi:}
\providecommand{\ArXivprefix}{arXiv:}
\providecommand{\URLprefix}{URL: }
\providecommand{\Pubmedprefix}{pmid:}
\providecommand{\doi}[1]{\href{http://dx.doi.org/#1}{\path{#1}}}
\providecommand{\Pubmed}[1]{\href{pmid:#1}{\path{#1}}}
\providecommand{\bibinfo}[2]{#2}
\ifx\xfnm\relax \def\xfnm[#1]{\unskip,\space#1}\fi
\bibitem[{Agga et~al.(2021)Agga, Abbou, Labbadi and El~Houm}]{agga2021short}
\bibinfo{author}{Agga, A.}, \bibinfo{author}{Abbou, A.},
  \bibinfo{author}{Labbadi, M.}, \bibinfo{author}{El~Houm, Y.},
  \bibinfo{year}{2021}.
\newblock \bibinfo{title}{Short-term self consumption {PV} plant power
  production forecasts based on hybrid {CNN-LSTM}, {ConvLSTM} models}.
\newblock \bibinfo{journal}{Renewable Energy} \bibinfo{volume}{177},
  \bibinfo{pages}{101--112}.
\bibitem[{Alexandrov et~al.(2020)Alexandrov, Benidis, Bohlke-Schneider,
  Flunkert, Gasthaus, Januschowski, Maddix, Rangapuram, Salinas, Schulz
  et~al.}]{alexandrov2020gluonts}
\bibinfo{author}{Alexandrov, A.}, \bibinfo{author}{Benidis, K.},
  \bibinfo{author}{Bohlke-Schneider, M.}, \bibinfo{author}{Flunkert, V.},
  \bibinfo{author}{Gasthaus, J.}, \bibinfo{author}{Januschowski, T.},
  \bibinfo{author}{Maddix, D.C.}, \bibinfo{author}{Rangapuram, S.},
  \bibinfo{author}{Salinas, D.}, \bibinfo{author}{Schulz, J.}, et~al.,
  \bibinfo{year}{2020}.
\newblock \bibinfo{title}{{GluonTS}: Probabilistic and neural time series
  modeling in python}.
\newblock \bibinfo{journal}{Journal of Machine Learning Research}
  \bibinfo{volume}{21}, \bibinfo{pages}{1--6}.
\bibitem[{Alippi et~al.(2018)Alippi, Disabato and Roveri}]{alippi2018moving}
\bibinfo{author}{Alippi, C.}, \bibinfo{author}{Disabato, S.},
  \bibinfo{author}{Roveri, M.}, \bibinfo{year}{2018}.
\newblock \bibinfo{title}{Moving convolutional neural networks to embedded
  systems: the alexnet and {VGG-16} case}, in: \bibinfo{booktitle}{2018 17th
  ACM/IEEE International Conference on Information Processing in Sensor
  Networks (IPSN)}, \bibinfo{organization}{IEEE}. pp.
  \bibinfo{pages}{212--223}.
\bibitem[{Alzubaidi et~al.(2021)Alzubaidi, Zhang, Humaidi, Al-Dujaili, Duan,
  Al-Shamma, Santamar{\'\i}a, Fadhel, Al-Amidie and
  Farhan}]{alzubaidi2021review}
\bibinfo{author}{Alzubaidi, L.}, \bibinfo{author}{Zhang, J.},
  \bibinfo{author}{Humaidi, A.J.}, \bibinfo{author}{Al-Dujaili, A.},
  \bibinfo{author}{Duan, Y.}, \bibinfo{author}{Al-Shamma, O.},
  \bibinfo{author}{Santamar{\'\i}a, J.}, \bibinfo{author}{Fadhel, M.A.},
  \bibinfo{author}{Al-Amidie, M.}, \bibinfo{author}{Farhan, L.},
  \bibinfo{year}{2021}.
\newblock \bibinfo{title}{Review of deep learning: Concepts, cnn architectures,
  challenges, applications, future directions}.
\newblock \bibinfo{journal}{Journal of Big Data} \bibinfo{volume}{8},
  \bibinfo{pages}{1--74}.
\bibitem[{Ballester and Araujo(2016)}]{ballester2016performance}
\bibinfo{author}{Ballester, P.}, \bibinfo{author}{Araujo, R.M.},
  \bibinfo{year}{2016}.
\newblock \bibinfo{title}{On the performance of googlenet and alexnet applied
  to sketches}, in: \bibinfo{booktitle}{Thirtieth AAAI Conference on Artificial
  Intelligence}.
\bibitem[{Box et~al.(2015)Box, Jenkins, Reinsel and Ljung}]{box2015time}
\bibinfo{author}{Box, G.E.}, \bibinfo{author}{Jenkins, G.M.},
  \bibinfo{author}{Reinsel, G.C.}, \bibinfo{author}{Ljung, G.M.},
  \bibinfo{year}{2015}.
\newblock \bibinfo{title}{Time series analysis: forecasting and control}.
\newblock \bibinfo{publisher}{John Wiley \& Sons}.
\bibitem[{Canziani et~al.(2016)Canziani, Paszke and
  Culurciello}]{canziani2016analysis}
\bibinfo{author}{Canziani, A.}, \bibinfo{author}{Paszke, A.},
  \bibinfo{author}{Culurciello, E.}, \bibinfo{year}{2016}.
\newblock \bibinfo{title}{An analysis of deep neural network models for
  practical applications}.
\newblock \bibinfo{journal}{arXiv preprint arXiv:1605.07678} .
\bibitem[{Chen et~al.(2020a)Chen, Li, Zhang and Li}]{chen2020short}
\bibinfo{author}{Chen, G.}, \bibinfo{author}{Li, L.}, \bibinfo{author}{Zhang,
  Z.}, \bibinfo{author}{Li, S.}, \bibinfo{year}{2020}a.
\newblock \bibinfo{title}{Short-term wind speed forecasting with
  principle-subordinate predictor based on {Conv-LSTM} and improved {BPNN}}.
\newblock \bibinfo{journal}{IEEE Access} \bibinfo{volume}{8},
  \bibinfo{pages}{67955--67973}.
\bibitem[{Chen et~al.(2020b)Chen, Kang, Chen and Wang}]{chen2020probabilistic}
\bibinfo{author}{Chen, Y.}, \bibinfo{author}{Kang, Y.}, \bibinfo{author}{Chen,
  Y.}, \bibinfo{author}{Wang, Z.}, \bibinfo{year}{2020}b.
\newblock \bibinfo{title}{Probabilistic forecasting with temporal convolutional
  neural network}.
\newblock \bibinfo{journal}{Neurocomputing} .
\bibitem[{Chung et~al.(2014)Chung, Gulcehre, Cho and
  Bengio}]{chung2014empirical}
\bibinfo{author}{Chung, J.}, \bibinfo{author}{Gulcehre, C.},
  \bibinfo{author}{Cho, K.}, \bibinfo{author}{Bengio, Y.},
  \bibinfo{year}{2014}.
\newblock \bibinfo{title}{Empirical evaluation of gated recurrent neural
  networks on sequence modeling}.
\newblock \bibinfo{journal}{arXiv preprint arXiv:1412.3555} .
\bibitem[{Dehghani et~al.(2019)Dehghani, Riahi-Madvar, Hooshyaripor, Mosavi,
  Shamshirband, Zavadskas and Chau}]{dehghani2019prediction}
\bibinfo{author}{Dehghani, M.}, \bibinfo{author}{Riahi-Madvar, H.},
  \bibinfo{author}{Hooshyaripor, F.}, \bibinfo{author}{Mosavi, A.},
  \bibinfo{author}{Shamshirband, S.}, \bibinfo{author}{Zavadskas, E.K.},
  \bibinfo{author}{Chau, K.w.}, \bibinfo{year}{2019}.
\newblock \bibinfo{title}{Prediction of hydropower generation using grey wolf
  optimization adaptive neuro-fuzzy inference system}.
\newblock \bibinfo{journal}{Energies} \bibinfo{volume}{12},
  \bibinfo{pages}{289}.
\bibitem[{Dorado-Moreno et~al.(2020)Dorado-Moreno, Navarin, Guti{\'e}rrez,
  Prieto, Sperduti, Salcedo-Sanz and
  Herv{\'a}s-Mart{\'\i}nez}]{dorado2020multi}
\bibinfo{author}{Dorado-Moreno, M.}, \bibinfo{author}{Navarin, N.},
  \bibinfo{author}{Guti{\'e}rrez, P.A.}, \bibinfo{author}{Prieto, L.},
  \bibinfo{author}{Sperduti, A.}, \bibinfo{author}{Salcedo-Sanz, S.},
  \bibinfo{author}{Herv{\'a}s-Mart{\'\i}nez, C.}, \bibinfo{year}{2020}.
\newblock \bibinfo{title}{Multi-task learning for the prediction of wind power
  ramp events with deep neural networks}.
\newblock \bibinfo{journal}{Neural Networks} \bibinfo{volume}{123},
  \bibinfo{pages}{401--411}.
\bibitem[{Fawaz et~al.(2019)Fawaz, Forestier, Weber, Idoumghar and
  Muller}]{fawaz2019deep}
\bibinfo{author}{Fawaz, H.I.}, \bibinfo{author}{Forestier, G.},
  \bibinfo{author}{Weber, J.}, \bibinfo{author}{Idoumghar, L.},
  \bibinfo{author}{Muller, P.A.}, \bibinfo{year}{2019}.
\newblock \bibinfo{title}{Deep learning for time series classification: a
  review}.
\newblock \bibinfo{journal}{Data Mining and Knowledge Discovery}
  \bibinfo{volume}{33}, \bibinfo{pages}{917--963}.
\bibitem[{Galicia et~al.(2019)Galicia, Talavera-Llames, Troncoso, Koprinska and
  Martínez-Álvarez}]{galacia2019}
\bibinfo{author}{Galicia, A.}, \bibinfo{author}{Talavera-Llames, R.},
  \bibinfo{author}{Troncoso, A.}, \bibinfo{author}{Koprinska, I.},
  \bibinfo{author}{Martínez-Álvarez, F.}, \bibinfo{year}{2019}.
\newblock \bibinfo{title}{Multi-step forecasting for big data time series based
  on ensemble learning}.
\newblock \bibinfo{journal}{Knowledge-Based Systems} \bibinfo{volume}{163},
  \bibinfo{pages}{830--841}.
\bibitem[{Gao et~al.(2019)Gao, Li, Hong and Long}]{gao2019day}
\bibinfo{author}{Gao, M.}, \bibinfo{author}{Li, J.}, \bibinfo{author}{Hong,
  F.}, \bibinfo{author}{Long, D.}, \bibinfo{year}{2019}.
\newblock \bibinfo{title}{Day-ahead power forecasting in a large-scale
  photovoltaic plant based on weather classification using {LSTM}}.
\newblock \bibinfo{journal}{Energy} \bibinfo{volume}{187},
  \bibinfo{pages}{115838}.
\bibitem[{Grigsby et~al.(2021)Grigsby, Wang and Qi}]{grigsby2021long}
\bibinfo{author}{Grigsby, J.}, \bibinfo{author}{Wang, Z.}, \bibinfo{author}{Qi,
  Y.}, \bibinfo{year}{2021}.
\newblock \bibinfo{title}{Long-range transformers for dynamic spatiotemporal
  forecasting}.
\newblock \bibinfo{journal}{arXiv preprint arXiv:2109.12218} .
\bibitem[{Gu et~al.(2018)Gu, Wang, Kuen, Ma, Shahroudy, Shuai, Liu, Wang, Wang,
  Cai et~al.}]{gu2018recent}
\bibinfo{author}{Gu, J.}, \bibinfo{author}{Wang, Z.}, \bibinfo{author}{Kuen,
  J.}, \bibinfo{author}{Ma, L.}, \bibinfo{author}{Shahroudy, A.},
  \bibinfo{author}{Shuai, B.}, \bibinfo{author}{Liu, T.},
  \bibinfo{author}{Wang, X.}, \bibinfo{author}{Wang, G.}, \bibinfo{author}{Cai,
  J.}, et~al., \bibinfo{year}{2018}.
\newblock \bibinfo{title}{Recent advances in convolutional neural networks}.
\newblock \bibinfo{journal}{Pattern Recognition} \bibinfo{volume}{77},
  \bibinfo{pages}{354--377}.
\bibitem[{Hanifi et~al.(2020)Hanifi, Liu, Lin and Lotfian}]{hanifi2020critical}
\bibinfo{author}{Hanifi, S.}, \bibinfo{author}{Liu, X.}, \bibinfo{author}{Lin,
  Z.}, \bibinfo{author}{Lotfian, S.}, \bibinfo{year}{2020}.
\newblock \bibinfo{title}{A critical review of wind power forecasting
  methods—past, present and future}.
\newblock \bibinfo{journal}{Energies} \bibinfo{volume}{13},
  \bibinfo{pages}{3764}.
\bibitem[{Hewamalage et~al.(2022)Hewamalage, Bergmeir and
  Bandara}]{hewamalage2022global}
\bibinfo{author}{Hewamalage, H.}, \bibinfo{author}{Bergmeir, C.},
  \bibinfo{author}{Bandara, K.}, \bibinfo{year}{2022}.
\newblock \bibinfo{title}{Global models for time series forecasting: A
  simulation study}.
\newblock \bibinfo{journal}{Pattern Recognition} \bibinfo{volume}{124},
  \bibinfo{pages}{108441}.
\bibitem[{Hochreiter and Schmidhuber(1997)}]{hochreiter1997long}
\bibinfo{author}{Hochreiter, S.}, \bibinfo{author}{Schmidhuber, J.},
  \bibinfo{year}{1997}.
\newblock \bibinfo{title}{Long short-term memory}.
\newblock \bibinfo{journal}{Neural Computation} \bibinfo{volume}{9},
  \bibinfo{pages}{1735--1780}.
\bibitem[{Hsu and Lachenbruch(2014)}]{hsu2014paired}
\bibinfo{author}{Hsu, H.}, \bibinfo{author}{Lachenbruch, P.},
  \bibinfo{year}{2014}.
\newblock \bibinfo{title}{Paired t test}.
\newblock \bibinfo{journal}{Wiley StatsRef: Statistics Reference Online} .
\bibitem[{Iandola et~al.(2016)Iandola, Han, Moskewicz, Ashraf, Dally and
  Keutzer}]{iandola2016squeezenet}
\bibinfo{author}{Iandola, F.N.}, \bibinfo{author}{Han, S.},
  \bibinfo{author}{Moskewicz, M.W.}, \bibinfo{author}{Ashraf, K.},
  \bibinfo{author}{Dally, W.J.}, \bibinfo{author}{Keutzer, K.},
  \bibinfo{year}{2016}.
\newblock \bibinfo{title}{{SqueezeNet}: {AlexNet}-level accuracy with $50x$
  fewer parameters and $< 0.5$ {MB} model size}.
\newblock \bibinfo{journal}{arXiv preprint arXiv:1602.07360} .
\bibitem[{Ilic et~al.(2020)Ilic, Gorgulu and Cevik}]{ilic2020augmented}
\bibinfo{author}{Ilic, I.}, \bibinfo{author}{Gorgulu, B.},
  \bibinfo{author}{Cevik, M.}, \bibinfo{year}{2020}.
\newblock \bibinfo{title}{Augmented out-of-sample comparison method for time
  series forecasting techniques}, in: \bibinfo{booktitle}{Canadian Conference
  on Artificial Intelligence}, \bibinfo{organization}{Springer}. pp.
  \bibinfo{pages}{302--308}.
\bibitem[{Ilic et~al.(2021)Ilic, G{\"o}rg{\"u}l{\"u}, Cevik and
  Baydo{\u{g}}an}]{ilic2020b}
\bibinfo{author}{Ilic, I.}, \bibinfo{author}{G{\"o}rg{\"u}l{\"u}, B.},
  \bibinfo{author}{Cevik, M.}, \bibinfo{author}{Baydo{\u{g}}an, M.G.},
  \bibinfo{year}{2021}.
\newblock \bibinfo{title}{Explainable boosted linear regression for time series
  forecasting}.
\newblock \bibinfo{journal}{Pattern Recognition} \bibinfo{volume}{120},
  \bibinfo{pages}{108144}.
\bibitem[{Jiajun et~al.(2020)Jiajun, Chuanjin, Yongle and
  Huoyue}]{jiajun2020ultra}
\bibinfo{author}{Jiajun, H.}, \bibinfo{author}{Chuanjin, Y.},
  \bibinfo{author}{Yongle, L.}, \bibinfo{author}{Huoyue, X.},
  \bibinfo{year}{2020}.
\newblock \bibinfo{title}{Ultra-short term wind prediction with wavelet
  transform, deep belief network and ensemble learning}.
\newblock \bibinfo{journal}{Energy Conversion and Management}
  \bibinfo{volume}{205}, \bibinfo{pages}{112418}.
\bibitem[{Ju et~al.(2019)Ju, Sun, Chen, Zhang, Zhu and Rehman}]{ju2019model}
\bibinfo{author}{Ju, Y.}, \bibinfo{author}{Sun, G.}, \bibinfo{author}{Chen,
  Q.}, \bibinfo{author}{Zhang, M.}, \bibinfo{author}{Zhu, H.},
  \bibinfo{author}{Rehman, M.U.}, \bibinfo{year}{2019}.
\newblock \bibinfo{title}{A model combining convolutional neural network and
  {LightGBM} algorithm for ultra-short-term wind power forecasting}.
\newblock \bibinfo{journal}{{IEEE Access}} \bibinfo{volume}{7},
  \bibinfo{pages}{28309--28318}.
\bibitem[{Kazutoshi et~al.(2018)Kazutoshi, Yu and
  Yasuhiro}]{kazutoshi2018feature}
\bibinfo{author}{Kazutoshi, H.}, \bibinfo{author}{Yu, F.},
  \bibinfo{author}{Yasuhiro, H.}, \bibinfo{year}{2018}.
\newblock \bibinfo{title}{Feature extraction of nwp data for wind power
  forecasting using 3d-convolutional neural networks-sciencedirect}.
\newblock \bibinfo{journal}{Energy Procedia} \bibinfo{volume}{155},
  \bibinfo{pages}{350--358}.
\bibitem[{Ke et~al.(2017)Ke, Meng, Finley, Wang, Chen, Ma, Ye and
  Liu}]{ke2017lightgbm}
\bibinfo{author}{Ke, G.}, \bibinfo{author}{Meng, Q.}, \bibinfo{author}{Finley,
  T.}, \bibinfo{author}{Wang, T.}, \bibinfo{author}{Chen, W.},
  \bibinfo{author}{Ma, W.}, \bibinfo{author}{Ye, Q.}, \bibinfo{author}{Liu,
  T.Y.}, \bibinfo{year}{2017}.
\newblock \bibinfo{title}{{LightGBM}: A highly efficient gradient boosting
  decision tree}.
\newblock \bibinfo{journal}{Advances in Neural Information Processing Systems}
  \bibinfo{volume}{30}, \bibinfo{pages}{3146--3154}.
\bibitem[{Khosravi et~al.(2022)Khosravi, Golkarian and
  Tiefenbacher}]{khosravi2022using}
\bibinfo{author}{Khosravi, K.}, \bibinfo{author}{Golkarian, A.},
  \bibinfo{author}{Tiefenbacher, J.P.}, \bibinfo{year}{2022}.
\newblock \bibinfo{title}{Using optimized deep learning to predict daily
  streamflow: A comparison to common machine learning algorithms}.
\newblock \bibinfo{journal}{Water Resources Management} \bibinfo{volume}{36},
  \bibinfo{pages}{699--716}.
\bibitem[{Lim et~al.(2021)Lim, Ar{\i}k, Loeff and Pfister}]{lim2021temporal}
\bibinfo{author}{Lim, B.}, \bibinfo{author}{Ar{\i}k, S.{\"O}.},
  \bibinfo{author}{Loeff, N.}, \bibinfo{author}{Pfister, T.},
  \bibinfo{year}{2021}.
\newblock \bibinfo{title}{Temporal fusion transformers for interpretable
  multi-horizon time series forecasting}.
\newblock \bibinfo{journal}{International Journal of Forecasting}
  \bibinfo{volume}{37}, \bibinfo{pages}{1748--1764}.
\bibitem[{Liu et~al.(2017)Liu, Wang, Liu and Zhu}]{liu2017towards}
\bibinfo{author}{Liu, S.}, \bibinfo{author}{Wang, X.}, \bibinfo{author}{Liu,
  M.}, \bibinfo{author}{Zhu, J.}, \bibinfo{year}{2017}.
\newblock \bibinfo{title}{Towards better analysis of machine learning models: A
  visual analytics perspective}.
\newblock \bibinfo{journal}{Visual Informatics} \bibinfo{volume}{1},
  \bibinfo{pages}{48--56}.
\bibitem[{Makridakis et~al.(2020)Makridakis, Spiliotis and
  Assimakopoulos}]{makridakis2020m4}
\bibinfo{author}{Makridakis, S.}, \bibinfo{author}{Spiliotis, E.},
  \bibinfo{author}{Assimakopoulos, V.}, \bibinfo{year}{2020}.
\newblock \bibinfo{title}{The m4 competition: 100,000 time series and 61
  forecasting methods}.
\newblock \bibinfo{journal}{International Journal of Forecasting}
  \bibinfo{volume}{36}, \bibinfo{pages}{54--74}.
\bibitem[{Mirjalili et~al.(2014)Mirjalili, Mirjalili and
  Lewis}]{mirjalili2014grey}
\bibinfo{author}{Mirjalili, S.}, \bibinfo{author}{Mirjalili, S.M.},
  \bibinfo{author}{Lewis, A.}, \bibinfo{year}{2014}.
\newblock \bibinfo{title}{Grey wolf optimizer}.
\newblock \bibinfo{journal}{Advances in Engineering Software}
  \bibinfo{volume}{69}, \bibinfo{pages}{46--61}.
\bibitem[{Montero-Manso and Hyndman(2021)}]{montero2021principles}
\bibinfo{author}{Montero-Manso, P.}, \bibinfo{author}{Hyndman, R.J.},
  \bibinfo{year}{2021}.
\newblock \bibinfo{title}{Principles and algorithms for forecasting groups of
  time series: Locality and globality}.
\newblock \bibinfo{journal}{International Journal of Forecasting}
  \bibinfo{volume}{37}, \bibinfo{pages}{1632--1653}.
\bibitem[{Oreshkin et~al.(2019)Oreshkin, Carpov, Chapados and
  Bengio}]{oreshkin2019n}
\bibinfo{author}{Oreshkin, B.N.}, \bibinfo{author}{Carpov, D.},
  \bibinfo{author}{Chapados, N.}, \bibinfo{author}{Bengio, Y.},
  \bibinfo{year}{2019}.
\newblock \bibinfo{title}{{N-BEATS}: Neural basis expansion analysis for
  interpretable time series forecasting}.
\newblock \bibinfo{journal}{arXiv preprint arXiv:1905.10437} .
\bibitem[{Ozoegwu(2019)}]{ozoegwu2019artificial}
\bibinfo{author}{Ozoegwu, C.G.}, \bibinfo{year}{2019}.
\newblock \bibinfo{title}{Artificial neural network forecast of monthly mean
  daily global solar radiation of selected locations based on time series and
  month number}.
\newblock \bibinfo{journal}{Journal of Cleaner Production}
  \bibinfo{volume}{216}, \bibinfo{pages}{1--13}.
\bibitem[{Parmezan et~al.(2019)Parmezan, Souza and Batista}]{parmezan2019}
\bibinfo{author}{Parmezan, A.R.S.}, \bibinfo{author}{Souza, V.M.},
  \bibinfo{author}{Batista, G.E.}, \bibinfo{year}{2019}.
\newblock \bibinfo{title}{Evaluation of statistical and machine learning models
  for time series prediction: Identifying the state-of-the-art and the best
  conditions for the use of each model}.
\newblock \bibinfo{journal}{Information Sciences} \bibinfo{volume}{484},
  \bibinfo{pages}{302--337}.
\bibitem[{Rangapuram et~al.(2018)Rangapuram, Seeger, Gasthaus, Stella, Wang and
  Januschowski}]{rangapuram2018deep}
\bibinfo{author}{Rangapuram, S.S.}, \bibinfo{author}{Seeger, M.W.},
  \bibinfo{author}{Gasthaus, J.}, \bibinfo{author}{Stella, L.},
  \bibinfo{author}{Wang, Y.}, \bibinfo{author}{Januschowski, T.},
  \bibinfo{year}{2018}.
\newblock \bibinfo{title}{Deep state space models for time series forecasting},
  in: \bibinfo{booktitle}{Advances in Neural Information Processing Systems},
  pp. \bibinfo{pages}{7785--7794}.
\bibitem[{Salinas et~al.(2020)Salinas, Flunkert, Gasthaus and
  Januschowski}]{salinas2020deepar}
\bibinfo{author}{Salinas, D.}, \bibinfo{author}{Flunkert, V.},
  \bibinfo{author}{Gasthaus, J.}, \bibinfo{author}{Januschowski, T.},
  \bibinfo{year}{2020}.
\newblock \bibinfo{title}{{DeepAR}: Probabilistic forecasting with
  autoregressive recurrent networks}.
\newblock \bibinfo{journal}{International Journal of Forecasting}
  \bibinfo{volume}{36}, \bibinfo{pages}{1181--1191}.
\bibitem[{Sharma et~al.(2011)Sharma, Sharma, Irwin and
  Shenoy}]{sharma2011predicting}
\bibinfo{author}{Sharma, N.}, \bibinfo{author}{Sharma, P.},
  \bibinfo{author}{Irwin, D.}, \bibinfo{author}{Shenoy, P.},
  \bibinfo{year}{2011}.
\newblock \bibinfo{title}{Predicting solar generation from weather forecasts
  using machine learning}, in: \bibinfo{booktitle}{2011 IEEE International
  Conference on Smart Grid Communications (SmartGridComm)},
  \bibinfo{organization}{IEEE}. pp. \bibinfo{pages}{528--533}.
\bibitem[{Shi et~al.(2015)Shi, Chen, Wang, Yeung, Wong and
  Woo}]{shi2015convolutional}
\bibinfo{author}{Shi, X.}, \bibinfo{author}{Chen, Z.}, \bibinfo{author}{Wang,
  H.}, \bibinfo{author}{Yeung, D.Y.}, \bibinfo{author}{Wong, W.K.},
  \bibinfo{author}{Woo, W.c.}, \bibinfo{year}{2015}.
\newblock \bibinfo{title}{Convolutional {LSTM} network: A machine learning
  approach for precipitation nowcasting}.
\newblock \bibinfo{journal}{Advances in Neural Information Processing Systems}
  \bibinfo{volume}{28}.
\bibitem[{Taherkhani et~al.(2020)Taherkhani, Cosma and
  McGinnity}]{taherkhani2020adaboost}
\bibinfo{author}{Taherkhani, A.}, \bibinfo{author}{Cosma, G.},
  \bibinfo{author}{McGinnity, T.M.}, \bibinfo{year}{2020}.
\newblock \bibinfo{title}{{AdaBoost-CNN}: An adaptive boosting algorithm for
  convolutional neural networks to classify multi-class imbalanced datasets
  using transfer learning}.
\newblock \bibinfo{journal}{Neurocomputing} \bibinfo{volume}{404},
  \bibinfo{pages}{351--366}.
\bibitem[{Voyant et~al.(2017)Voyant, Notton, Kalogirou, Nivet, Paoli, Motte and
  Fouilloy}]{voyant2017machine}
\bibinfo{author}{Voyant, C.}, \bibinfo{author}{Notton, G.},
  \bibinfo{author}{Kalogirou, S.}, \bibinfo{author}{Nivet, M.L.},
  \bibinfo{author}{Paoli, C.}, \bibinfo{author}{Motte, F.},
  \bibinfo{author}{Fouilloy, A.}, \bibinfo{year}{2017}.
\newblock \bibinfo{title}{Machine learning methods for solar radiation
  forecasting: A review}.
\newblock \bibinfo{journal}{Renewable Energy} \bibinfo{volume}{105},
  \bibinfo{pages}{569--582}.
\bibitem[{Wu et~al.(2021)Wu, Guan, Lv and Huang}]{wu2021ultra}
\bibinfo{author}{Wu, Q.}, \bibinfo{author}{Guan, F.}, \bibinfo{author}{Lv, C.},
  \bibinfo{author}{Huang, Y.}, \bibinfo{year}{2021}.
\newblock \bibinfo{title}{Ultra-short-term multi-step wind power forecasting
  based on {CNN-LSTM}}.
\newblock \bibinfo{journal}{IET Renewable Power Generation}
  \bibinfo{volume}{15}, \bibinfo{pages}{1019--1029}.
\bibitem[{Xiang et~al.(2022)Xiang, Liu, Yang, Hu and Su}]{xiang2022ultra}
\bibinfo{author}{Xiang, L.}, \bibinfo{author}{Liu, J.}, \bibinfo{author}{Yang,
  X.}, \bibinfo{author}{Hu, A.}, \bibinfo{author}{Su, H.},
  \bibinfo{year}{2022}.
\newblock \bibinfo{title}{Ultra-short term wind power prediction applying a
  novel model named {SATCN-LSTM}}.
\newblock \bibinfo{journal}{Energy Conversion and Management}
  \bibinfo{volume}{252}, \bibinfo{pages}{115036}.
\bibitem[{Xiong et~al.(2022)Xiong, Lou, Meng, Wang, Ma and
  Wang}]{xiong2022short}
\bibinfo{author}{Xiong, B.}, \bibinfo{author}{Lou, L.}, \bibinfo{author}{Meng,
  X.}, \bibinfo{author}{Wang, X.}, \bibinfo{author}{Ma, H.},
  \bibinfo{author}{Wang, Z.}, \bibinfo{year}{2022}.
\newblock \bibinfo{title}{Short-term wind power forecasting based on attention
  mechanism and deep learning}.
\newblock \bibinfo{journal}{Electric Power Systems Research}
  \bibinfo{volume}{206}, \bibinfo{pages}{107776}.
\bibitem[{Yang et~al.(2018)Yang, Wu, Qiu, Wang and Chen}]{yang2018emotion}
\bibinfo{author}{Yang, Y.}, \bibinfo{author}{Wu, Q.}, \bibinfo{author}{Qiu,
  M.}, \bibinfo{author}{Wang, Y.}, \bibinfo{author}{Chen, X.},
  \bibinfo{year}{2018}.
\newblock \bibinfo{title}{Emotion recognition from multi-channel {EEG} through
  parallel convolutional recurrent neural network}, in:
  \bibinfo{booktitle}{2018 International Joint Conference on Neural Networks
  (IJCNN)}, \bibinfo{organization}{IEEE}. pp. \bibinfo{pages}{1--7}.
\bibitem[{Yildiz et~al.(2021)Yildiz, Acikgoz, Korkmaz and
  Budak}]{yildiz2021improved}
\bibinfo{author}{Yildiz, C.}, \bibinfo{author}{Acikgoz, H.},
  \bibinfo{author}{Korkmaz, D.}, \bibinfo{author}{Budak, U.},
  \bibinfo{year}{2021}.
\newblock \bibinfo{title}{An improved residual-based convolutional neural
  network for very short-term wind power forecasting}.
\newblock \bibinfo{journal}{Energy Conversion and Management}
  \bibinfo{volume}{228}, \bibinfo{pages}{113731}.
\bibitem[{Zhen et~al.(2020)Zhen, Niu, Yu, Wang, Liang and Xu}]{zhen2020hybrid}
\bibinfo{author}{Zhen, H.}, \bibinfo{author}{Niu, D.}, \bibinfo{author}{Yu,
  M.}, \bibinfo{author}{Wang, K.}, \bibinfo{author}{Liang, Y.},
  \bibinfo{author}{Xu, X.}, \bibinfo{year}{2020}.
\newblock \bibinfo{title}{A hybrid deep learning model and comparison for wind
  power forecasting considering temporal-spatial feature extraction}.
\newblock \bibinfo{journal}{Sustainability} \bibinfo{volume}{12},
  \bibinfo{pages}{9490}.

\end{thebibliography}

\end{document}